\documentclass[10pt,journal,compsoc]{IEEEtran}
\usepackage{times}
\usepackage{epsfig}
\usepackage{graphicx}
\usepackage{amsmath}
\usepackage{amssymb}
\usepackage{subcaption}
\usepackage{titling}
\usepackage{float}
\usepackage{mathtools}
\usepackage{xcolor}
\usepackage[font=footnotesize]{caption}

\usepackage[ruled]{algorithm2e}
\usepackage{algorithmic}

\usepackage{booktabs}
\usepackage{multirow}

\usepackage[pagebackref=true,breaklinks=true,colorlinks,bookmarks=false]{hyperref}

\def\cvprPaperID{7985} 
\def\confYear{CVPR 2021}
\newtheorem{definition}{Definition}

\newcommand{\mc}[1]{\mathcal{#1}}
\newcommand{\mb}[1]{\mathbf{#1}}
\newcommand{\mr}[1]{\mathrm{#1}}
\newcommand{\mbb}[1]{\mathbb{#1}}
\newcommand{\bs}[1]{\boldsymbol{#1}}
\newcommand{\bx}{\mb{x}}
\newcommand{\by}{\mb{y}}
\newcommand{\bw}{\mb{w}}
\newcommand{\bW}{\mb{W}}
\newcommand{\bz}{\mb{z}}
\newcommand{\bv}{\mb{v}}
\newcommand{\bu}{\mb{u}}
\newcommand{\bh}{\mb{h}}
\newcommand{\bmu}{\bs{\mu}}
\newcommand{\bsg}{\bs{\sigma}}
\newcommand{\bepsilon}{\bs{\epsilon}}
\newcommand{\btheta}{\bs{\theta}}
\newcommand{\bpi}{\bs{\pi}}
\newcommand{\bphi}{\bs{\phi}}
\newcommand{\bbeta}{\bs{\beta}}
\newcommand{\KL}{\mathrm{KL}}
\newcommand{\calV}{{\cal V}}

\newcommand{\BNNN}{$\mr{BN}^3\,$}
\newcommand{\FNNN}{$\mr{FN}^3\,$}
\newcommand{\CUT}[1]{}

\newcommand{\abc}[1]{\textcolor{black}{#1}}

\usepackage{paralist}

\newcommand{\yedit}[1]{\textcolor{black}{#1}}

\begin{document}
\urlstyle{rm}

\pagenumbering{gobble}

\title{Variational Nested Dropout}

\author{Yufei Cui$^{1,2}$, Yu Mao$^1$, Ziquan Liu$^1$, Qiao Li$^3$, Antoni B. Chan$^1$, Xue Liu$^2$, Tei-Wei Kuo$^4$, Chun Jason Xue$^1$ \\
{\tt\small yufeicui92@gmail.com, yfcui@ibingli.com}\\
$^1$Department of Computer Science, City University of Hong Kong\\
$^2$School of Computer Science, McGill University\\
$^3$School of Informatics, Xiamen University\\
$^4$Department of Computer Science and Information Engineering, National Taiwan University
}

\date{}
\maketitle

\begin{abstract}
    Nested dropout~{\let\thefootnote\relax\footnote{\\Extension of \emph{Bayesian Nested Neural Networks for Uncertainty Calibration and Adaptive Compression}, CVPR21. \\Under review as a journal paper.}} is a variant of dropout operation that is able to order network parameters or features based on the pre-defined importance during training.
    It has been explored for: \textbf{I.} \emph{Constructing nested nets}~\cite{ijcai2020-288,Cui_2021_CVPR}: the nested nets are neural networks whose architectures can be adjusted instantly during testing time, e.g., based on computational constraints. 
    The nested dropout implicitly ranks the network parameters, generating a set of sub-networks such that any smaller sub-network forms the basis of a larger one.
    \textbf{II.} \emph{Learning ordered representation}~\cite{rippel2014learning}: the nested dropout applied to the latent representation of a generative model (e.g., auto-encoder) ranks the features, enforcing explicit order of the dense representation over dimensions.

    However, the dropout rate is fixed as a hyper-parameter during the whole training process. 
    For nested nets, when network parameters are removed, the performance decays in a human-specified trajectory rather than in a trajectory learned from data.
    For generative models, the importance of features is specified as a constant vector, restraining the flexibility of representation learning.
    To address the problem, we focus on the probabilistic counterpart of the nested dropout.
    We propose a variational nested dropout (VND) operation that draws samples of multi-dimensional ordered masks at a low cost, providing useful gradients to the parameters of nested dropout.
    Based on this approach, we design a Bayesian nested neural network that learns the order knowledge of the parameter distributions.
    We further exploit the VND under different generative models for learning ordered latent distributions.
    In experiments, we show that the proposed approach outperforms the nested network in terms of accuracy, calibration, and out-of-domain detection in classification tasks.
    It also outperforms the related generative models on data generation tasks.
\end{abstract}

\section{Introduction}
Modern deep neural networks (DNNs) have achieved great success in fields of supervised learning and representation learning.
In the meantime, deep learning models have a high demand for learning ordered information from data, for both \emph{model architecture} and \emph{representations}.

\noindent\textbf{\emph{Model architecture:}}
Deep learning models are experiencing rapid growth in model size and computation cost, which makes it difficult to deploy on diverse hardware platforms.
Recent works study how to develop a network with flexible size during test time~\cite{kim2018nestednet, yu2018slimmable, yu2019universally, cai2019once, ijcai2020-288, xu2020one}, to reduce the cost in designing~\cite{tan2019efficientnet}, training~\cite{kingma2014adam}, compressing~\cite{han2015deep} and deploying~\cite{ren2019admm} a DNN on various platforms.
As these networks are often composed of a nested set of smaller sub-networks, we refer to them as \emph{nested nets} in this paper.
In this set, any smaller sub-network forms the basis of a larger one, and the residual information is learned via the incremental parameters added to the smaller sub-network~\cite{ijcai2020-288}.
One basis for creating nested nets is to order the network components (e.g., convolution channels) such that less important components can be removed first when creating the sub-network.
To avoid significant performance drop when removing the components, training with explicitly \emph{ordering of the network components} is required.
%

%
%

\noindent\textbf{\emph{Representations:}}
Representation (feature) learning is a major branch for modern machine learning research.
The unsupervised feature learning is able to discover low-dimensional structure underlying the high-dimensional input data.
Typical examples with neural network include restricted Boltzmann machine~\cite{salakhutdinov2007restricted} and auto-encoder~\cite{vincent2010stacked}.
However, the standard approach only provides features that are entangled and with equivalent importance.
To clearly identify the more important features and reduce the redundancy, recent works study imposing structural constraints to \emph{order the learned representations}~\cite{rippel2014learning,bekasov2020ordering}.

An operator for neural networks, \emph{nested dropout} \abc{(also called \emph{ordered dropout})}, was proposed for both ordering in model architectures  \cite{ijcai2020-288,finn2014learning} and ordering representations \cite{rippel2014learning, bekasov2020ordering}.
It was first developed to order the latent feature representation for the encoder-decoder models~\cite{rippel2014learning, bekasov2020ordering}.
Specifically, a discrete distribution is assigned over the indices of the representations, 
and the operation of nested dropout samples an index then drops the representations with larger indices.
%
This imposes explicit ordered importance over \emph{representations} by different frequencies that different dimensions of latent representation are activated during training.

Recent studies show that the nested dropout is also able to order the \emph{network components} during training such that \emph{nested nets} can be obtained~\cite{ijcai2020-288,finn2014learning}.
The ordering layout is applicable to different granularity levels of network components: single weights, groups of weights, convolutional channels, residual blocks, network layers, and even quantization bits. 
We refer to the partitions of the network components as \emph{nodes} in this paper.

Despite its successes, nested dropout requires that the probability that an index is sampled is specified by hand as a hyper-parameter, and does not change during training.
Thus, the importance of 
nodes or representations are pre-determined by hand rather than learned from data.
To allow the dropout rate to be learned, we propose a fully Bayesian treatment for the nested dropout operator.
We first propose a new variant of the nested dropout, based on 
a chain of interdependent Bernoulli variables.
The chain simulates the Bernoulli trials and can be understood as a special case of a two-state Markov chain, which intuitively generates order information.
To save the time cost for sampling during training, we propose a new \emph{Downhill} distribution that approximates this chain.
This approximate posterior is built on Gumbel Softmax~\cite{jang2016categorical, maddison2016concrete}, which efficiently generates more flexible samples compared with the Bernoulli chain.
%
This allows efficient sampling of the multivariate \emph{ordered mask}, and provides useful gradients to update the importance of the nodes or representations. 
We name this major scheme as variational nested dropout (VND), 
\abc{and apply it to both ordering network components and ordering latent representations.}

First, based on the proposed VND, a Bayesian Nested Neural Network (\BNNN) is constructed with \emph{learnable ordered importance of network components},
where the 
independent distributions of nodes are interconnected with the ordering units.
%
A mixture model prior is placed over each node, while the model selection is determined by the Downhill samples (Figure~\ref{fig:overview}).
A variational inference problem is formulated and resolved, and 
we propose several methods to simplify the sampling and calculation of the regularization term.
The formulation is proved to be an generalization of ordered L0-regularization over the sub-networks.
The trained \BNNN is (sub-)optimal in the trade-off of model size and prediction performance.
The full Bayesian treatment further enhances the uncertainty calibration and out-of-domain detection ability of the nested nets.
To further exploit the performance gain from the VND, we proposed a two-step knowledge distillation framework to obtain high-performance deterministic student nested nets.

Second, we study using VND with the variational auto-encoder~\cite{kingma2013auto} to \emph{order the distributions over the latent representation}.
The diversity of generated data is enhanced due to the mixture nature of the latent distribution organized by VND.
Our design of approximate posterior guarantee the diversity for data generation.
We further applying VND to encode aleatoric uncertainty in the probabilistic UNet~\cite{kohl2018probabilistic}, which is used for applications whose labels are provided by multiple annotators.
The diversity of generated samples are enhanced by VND, and thus it captures the disagreement of noisy labels better.

Experiments on \BNNN with VND show that it outperforms the deterministic nested models with nested dropout in any sub-network, in terms of classification accuracy, calibration and out-of-domain detection.
The student nested nets obtained by knowledge distillation further improves the performance.
Results on variational auto-encoder with VND show it outperforms the baseline and related methods by a large margin in terms of image reconstruction and generation diversity.
\BNNN and probabilistic UNet with VND also outperform the vanilla probabilistic UNet~\cite{kohl2018probabilistic} on uncertainty-critical tasks with noisy labels.

In summary, the contributions of this paper are:
\begin{compactitem}
\item We propose a variational nested dropout (VND) unit with a novel pair of prior and posterior distributions.
\item We propose a novel Bayesian nested neural network (\BNNN) that can  generate large sets of uncertainty-calibrated sub-networks. The formulation can be viewed as a generalization of \emph{ordered} $\ell_0$-regularization over the sub-networks. We propose a distillation methods with VND for higher performance gain. 
\item We propose a new generative model, VND enhanced auto-encoder (VND-AE), whose training objective is shown to encourage diversity. We propose a new formulation using VND to encode the aleotoric uncertainty for capturing the disagreement in annotations.
\end{compactitem}

A preliminary version of this paper appears in \cite{Cui_2021_CVPR}. The major differences between this paper and \cite{Cui_2021_CVPR} are: 
\begin{itemize}
    \item \yedit{This paper generalizes the VND to the case of generative models, formulating a new family of generative models with an explicit ordered latent structure.
    The concrete implementations with VND include a variational autoencoder (VND-AE) and a probabilistic UNet (VND-PUNet).
    Experimental results show a highly competitive performance of our generative models in generating images, compared with the recent advances.}
    \item \yedit{This paper extends the \BNNN with a knowledge distillation (KD) process for higher performance gain from VND. The KD process maintains a nested structure for the generated student nets and is executed in a compatible way with the sampling of VND.}
\end{itemize}
    
The code\footnote{\url{https://github.com/ralphc1212/variational_nested_dropout}} and supplemental\footnote{\url{http://visal.cs.cityu.edu.hk/static/pubs/journal/pami-vnd-supp.pdf}} are publicly available.

The remainder of this paper is organized as follows.
\yedit{Section~\ref{sec:vnd} introduces the general formulation of variational nested dropout (VND), including the chain of Bernoulli variables as prior and the proposed Downhill distribution as the approximate posterior.
Section~\ref{sec:bnnn} provides the formulation of \BNNN using VND with a full Bayesian treatment.
Section~\ref{sec:vnd_generative} presents the probabilistic generative models with the ordered latent structure.
Section~\ref{sec:related} reviews the background and recent advances of this paper.
Section~\ref{sec:exp} shows the detailed evaluations of the proposed techniques.
Section~\ref{sec:conclusion} summarize the paper and provides possible future directions.
}


\section{Variational Nested Dropout}
\label{sec:vnd}
We first review nested dropout, and then propose our Bayesian ordering unit and variational approximation.

\subsection{A Review of Nested Dropout}
The previous works~\cite{rippel2014learning} that order the representations use either Geometric or Categorical distributions to sample the last index of the kept units, then drop the neurons with indices greater than it.
%
Specifically, the distribution $p_\mathbb{I}(\cdot)$ is assigned over the representation indices $1, \dots, K$.
The nested/ordered dropout operation proceeds as follows:
\begin{compactenum}
	\label{odo}
	\item \emph{Tail sampling:} A tail index is sampled $I\sim p_\mathbb{I}(\cdot)$ that represents the last element be kept. 
	\item \emph{Ordered dropping:} The elements with indices ${I+1}, \dots, K$ are dropped. 
\end{compactenum}
We also refer to this operation as an \emph{ordering unit} as the representations are sorted in order.

\begin{figure}[t]
	\begin{center}
		\includegraphics[width=\linewidth]{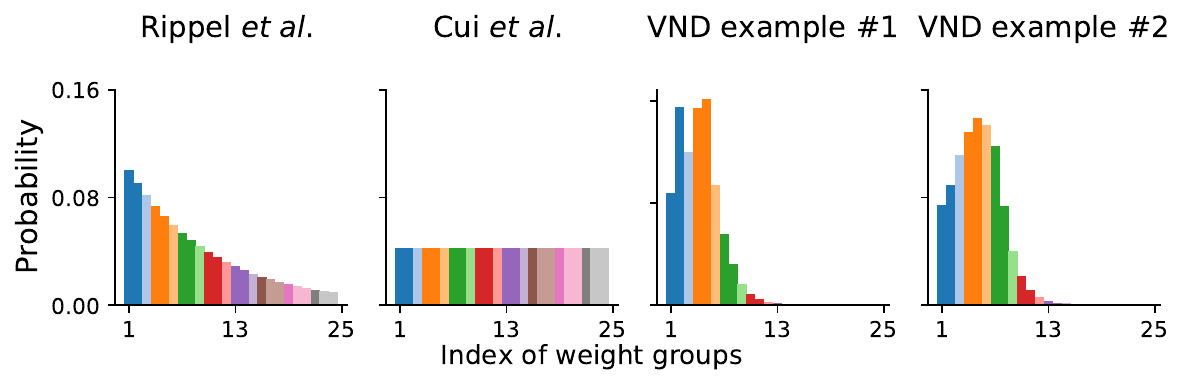}
	\end{center}
	\caption{The probability of tail index being sampled in different nested dropout realizations. Rippel~\emph{et al}~
	\cite{rippel2014learning} and Cui~\emph{et al}~
	\cite{ijcai2020-288} adopt Geometric and Categorical distributions, which are static over different layers and the learning process. The proposed variational nested dropout (VND) learns the importances of nodes or representations from data. The two examples are from two different layers in a Bayesian nested neural network.
	}
	\label{fig:motivation}
\end{figure}

In~\cite{rippel2014learning}, which focuses on learning ordered representations,  
this operation is proved to exactly recover PCA with a one-layer neural network.
%
%
%
Cui~\emph{et al}~\cite{ijcai2020-288} shows this operation, when applied to groups of neural network weights or quantization bits, generates nested sub-networks that are optimal for different computation resources.
They further prove that increasing from a smaller sub-network to a larger one maximizes the incremental information gain.
A large network only needs to be trained once with nested dropout, yielding a 
set of networks with varying sizes for deployment.
%
However, the above methods treat the nested dropout rate as a hyper-parameter, and hand-tuning the dropout rate is tedious and may lead to suboptimal performance of the sub-networks, as compared to learning this hyperparameter from the data.
%
%
As illustrated in Figure~\ref{fig:motivation}, the previous works use hand-specified parameters for nested dropout, which freezes the importance of the network components or representations over different layers during training.

A common practice for \emph{regular Bernoulli dropout} is to treat the dropout rate as a variational parameter in Bayesian neural networks~\cite{Gal2016Uncertainty}.
To find the optimal dropout rate, grid-search is first adopted~\cite{gal2016dropout}, whose complexity grows exponentially with the number of dropout units.
To alleviate the cost of searching, a continuous relaxation of the discrete dropout is proposed by which the dropout rate can be optimized directly~\cite{gal2017concrete},
improving accuracy and uncertainty, while keeping a low training time.
However, for \emph{nested dropout}, two aspects are unclear:  1) how to take a full Bayesian treatment with nested dropout units; 2) how the relaxation can be done for these units or how the gradients can be back-propagated to the parameters of $p_{\mbb{I}}(\cdot)$.
%
\subsection{Bayesian Ordering Unit}
\label{sec:bou}
The conventional nested dropout uses a Geometric distribution to sample the tail index $I$, $p_\mathbb{I}(I=i)=(1-\pi)^i \pi$, for $i\in\{1,\dots,K\}$.
By definition, the Geometric distribution models the probability that the $i$-th trial is the first ``success'' in a sequence of independent Bernoulli trials.
In the context of slimming neural networks, a ``failure'' of a Bernoulli trial indicates that node is kept, while a ``success'' indicates the tail index, where this node is kept and all subsequent nodes are dropped.
%
Thus, $\pi$ is the conditional probability of a node being a tail index, given the previous node is kept.
Sampling from the Geometric  only generates the tail index of the nodes to be kept.
A hard selection operation of ordered dropping is required to drop the following nodes.
The ordered dropping can be implemented using a set of ordered mask vectors $\calV = \{\mathbf{v}_1,\cdots,\mathbf{v}_K\}$, 
where $\mathbf{v}_j$ consists of $j$ ones followed by $K-j$ zeros, $\mathbf{v}_j =  [\underbrace{1,\dots,1}_{j},\underbrace{0,\dots,0}_{K-j}]$.
Given the sampled tail index $I\sim p_\mathbb{I}(\cdot)$, the appropriate mask $\mathbf{v}_I$ is selected and applied to the nodes (e.g., multiplying the weights).
However, as the masking is a non-differentiable transformation and does not provide a well-defined probability distribution, the nested dropout parameters cannot be learned using this formulation.

%

To find a more natural prior for the nodes, we propose to use a chain of Bernoulli variables 
to \emph{directly model the distribution of the ordered masks}.
%
Let the set of binary variables $\mathbf{z}=[z_1,\dots,z_K]$ represent the random ordered mask.
%
Specifically, we model the conditional distributions with Bernoulli variables,
\begin{align}
& p(z_1 = 1)=\pi_1,     & &p(z_1 = 0) =1-\pi_1,
\label{eq:berchain}\\
& p(z_i=1|z_{i-1}=1)=\pi_i, & &p(z_i=0|z_{i-1}=1)=1-\pi_i, 
\nonumber\\
& p(z_i=1|z_{i-1}=0)=0, & &p(z_i=0|z_{i-1}=0)=1,
\nonumber
\end{align}
where $\pi_i$ is the conditional probability of keeping the node given the previous node is kept, and $\pi_1=1$ (the first node is always kept).
Note that we also allow different probabilities $\pi_i$ for each $z_i$.
%
%
%
The marginal distribution of $z_i$ is
\begin{align}
 p(z_i=1)=\prod_{k=1}^i\pi_k, \quad p(z_i=0)=1-\prod_{k=1}^i\pi_k. \label{eq:marginal_z}
\end{align}
\yedit{We define the Bernoulli chain as $\rm{BernChain}(\bz,\bpi)$ with random variable $\bz$ and parameter $\bpi$.}
A property of this chain is that if $0$ occurs at the $i$-th position, the remaining elements with indices $i+1, \dots, K$ become $0$.
That is, sampling from this chain generates 
an ordered mask, which can be directly multiplied 
on the nodes to realize \emph{ordered dropping}.
Another benefit is that applying a continuous relaxation~\cite{gal2017concrete} of the Bernoulli variables in the chain allows its parameters $\boldsymbol{\pi}$
to be optimized.
However, the sampling of $\mathbf{z}$ requires stepping through each element $z_i$, which has complexity $\mathcal{O}(K)$, and is thus not scalable in modern DNNs where $K$ is large.
Thus we apply the variational inference framework, while \emph{treating $p(\mathbf{z})$ as the prior of the ordered mask} in our Bayesian treatment.
One challenge is to find a tractable variational distribution $q(\mathbf{z})$ that approximates the true posterior and is easy to compute.
%
%
Another challenge is to define a $q(\mathbf{z})$ that allows efficient re-parameterization, so that the gradient of the parameter of $q(\mathbf{z})$ can be estimated with low variance.
%
\subsection{Downhill Distribution as Approximate Posterior}
\label{sec:vou}

We next propose a novel \emph{Downhill} distribution based on Gumbel Softmax distribution~\cite{jang2016categorical, maddison2016concrete} that generates the ordered mask $\mathbf{z}$.
%
%
\yedit{Differently, the binary ordered mask with element $z_i$ is extended to the real values between range $[0,1]$.
Specifically, $z_i\in[0,1]$, $1\geq z_1\geq z_2\geq \dots \geq z_K\geq 0$.
}

\begin{definition}{Downhill Random Variables (r.v.).}
	Let the temperature parameter $\tau\in(0,\infty)$. An r.v. $\mathbf{z}$ has a Downhill distribution $\mathbf{z}\sim \mathrm{Downhill}(\bs{\beta},\tau)$, if its density is:
	\begin{align}
	&q(z_1,\dots,z_K)\\
	=&\Gamma(K)\tau^{K-1}\left[\sum_{i=1}^K\frac{ \beta_i}{(z_{i-1}-z_i)^\tau}\right]^{-K}\prod_{i=1}^K\frac{\beta_i}{(z_{i-1}-z_i)^{\tau+1}},\nonumber
	\end{align}
	where $\bs{\beta}=[\beta_1,\dots,\beta_K]$ 
	are the probabilities for each dimension.
\end{definition}

\begin{figure}[t]
	\begin{center}
		\includegraphics[width=\linewidth]{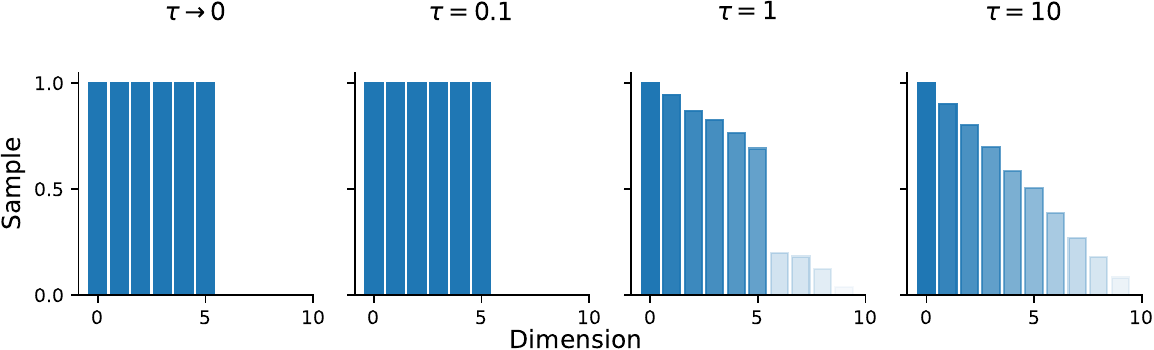}
	\end{center}
	\caption{The multivariate Downhill samples under different temperatures $\tau$. When $\tau\rightarrow0$, a clear cliff is observed as the dimension increases, which is beneficial for differentiating important or unimportant nodes. As $\tau$ increases, the shape becomes a slope where the gaps between important/unimportant nodes are smoother, which is beneficial for training.}
	\label{fig:sample_dh}
\end{figure}

\noindent
Two important properties of Downhill distributions are:

\begin{compactitem}
	\item {\bf Property 1.} If $\mathbf{c}\sim \mathrm{Gumbel\_softmax}(\tau,\beta,\epsilon_z)$\footnote{For Gumbel-softmax sampling, we first draw $g_1 \dots g_K$ from $\mathrm{Gumbel}(0,1)$, then calculate $c_i=\mr{softmax}(\frac{\log(\beta_i)+g_i}{\tau})$. The samples of $\mr{Gumbel}(0,1)$ can be obtained by first drawing $\epsilon_z\sim \mr{Uniform}(0,1)$ then computing $g=-\log(-\log(\epsilon_z))$.}, then $z_i=1-\mathrm{cumsum}_i^\prime(\mathbf{c})$, where $\mathbf{e}$ is a $K$-dimensional vector of ones, and $\mathrm{cumsum}_i^\prime(\mathbf{c})=\sum_{j=0}^{i-1}c_j$. $c_0\coloneqq1$. $\epsilon_z$ is a standard uniform variable.
	\item {\bf Property 2.} When $\tau\rightarrow 0$, sampling from the Downhill distribution reduces to discrete sampling, where the sample space is the set of ordered mask vectors, $\calV$. 
	The approximation of the Downhill distribution to the Bernoulli chain can be calculated in closed-form.
\end{compactitem}
Property 1 shows the sampling process of the Downhill distribution.
We visualize the Downhill samples in Figure~\ref{fig:sample_dh}.
As each multivariate sample has a shape of a long descent from left to right, we name it \emph{Downhill} distribution.
The temperature variable $\tau$ controls the sharpness of the downhill or the smoothness of the step at the tail index.
When $\tau$ is large, the slope is gentle in which case no nodes are dropped, but the less important nodes 
%
are multiplied with a factor less than 1.
When $\tau\rightarrow 0$, the shape of the sample becomes a cliff which is similar to the prior $p(\mathbf{z})$ on ordered masks,  
where the less important nodes are dropped (i.e., multiplied by 0).
Property 1 further implies the gradient $\frac{\partial}{\partial\bs{\beta}}\mbb{E}_{\bz\sim q_{\bs{\beta}}(\bz)}[\zeta(\bz)]$ can be estimated with low variance, for a cost function $\zeta(\bz)$.
Because the samples of $\bz$ are replaced by a differentiable function $t(\bs{\beta},\epsilon_z)$, then $\frac{\partial}{\partial\bs{\beta}}\mbb{E}_{\bz\sim q_{\bs{\beta}}(\bz)}[\zeta(\bz)]=\frac{\partial}{\partial\bs{\beta}}\mbb{E}_{\epsilon_z\sim \mr{Uniform}(0,1)}[\frac{\partial \zeta}{\partial t}\frac{\partial t}{\partial \bs{\beta}}]$, where $t(\cdot,\cdot)$ represents the whole transformation process in Prop.~1.

Recall that our objective is to approximate the chain of Bernoulli variables $p(\bz)$ with $q_{\bs{\beta}}(\bz)$.
Property 2 shows why the proposed distribution is consistent with the chain of Bernoullis in essence, and provides an easy way to derive the evidence lower bound (ELBO) for variational inference.
The proof for the two properties is in Appendix~\ref{ap-proof-property}.
This simple transformation of Gumbel softmax samples allows fast sampling of an ordering unit.
\yedit{Compared with $p(\mathbf{z})$, the only sequential process is a cumulative summation, which could be effectively accelerated by modern computation library.}
%

\section{Bayesian Nested Neural Network}
\label{sec:bnnn}
%
In this section, we present the Bayesian nested neural network (\BNNN) based on the fundamental units proposed in Section~\ref{sec:vnd}.
\setlength{\belowdisplayskip}{2pt} \setlength{\belowdisplayshortskip}{2pt}
\setlength{\abovedisplayskip}{2pt} \setlength{\abovedisplayshortskip}{2pt}

\subsection{Bayesian Inference and SGVB}
Consider a dataset $\mc{D}$ constructed from $N$ pairs of instances $\{(\bx_i, \by_i)\}_{i=1}^N$.
Our objective is to estimate the parameters $\bu$ of a neural network $p(\by|\bx, \bu)$ that predicts $\by$ given input $\bx$ and parameters $\bu$.
In Bayesian learning, a prior $p(\bu)$ is placed over the parameters $\bu$.
After data $\mc{D}$ is observed, the prior distribution is transformed into a posterior distribution $p(\bu|\mc{D})$.

For neural networks, 
computing the posterior distribution using the Bayes' rule requires computing intractable integrals over $\bu$.
%
Thus, approximation techniques are required.
One family of techniques is variational inference, with which the posterior $p(\bu|\mc{D})$ is approximated by a parametric distribution $q_{\bphi}(\bu)$, where $\bphi$ are the variational parameters.
$q_{\bphi}(\bu)$ is approximated by minimizing the Kullback-Leibler (KL) divergence with the true posterior,  $\KL[q_{\bphi}(\bu)||p(\bu|\mc{D})]$, which is equivalent to maximizing the \emph{evidence lower bound} (ELBO):
\begin{align}
	\mc{L}_{\bphi}
	=L_{\mc{D}}(\bphi) - \KL[q_{\bphi}(\bu)||p(\bu)]\label{eq:elbo1},
\end{align}
where the expected data log-likelihood is
\begin{equation}
	L_{\mc{D}}(\bphi)=\sum_{i=1}^N\mbb{E}_{q_{\bphi}(\bu)}[\log p(\by_i|\bx_i,\bu)].
\end{equation}
The integration $L_{\mc{D}}$ 
is not tractable for neural networks.
An efficient method for gradient-based optimization of the variational bound is \emph{stochastic gradient variational Bayes} (SGVB)~\cite{kingma2013auto, kingma2015variational}.
SGVB parameterizes the random variables (r.v.) $\bu\sim q_{\bphi}(\bu)$ as $\bu=t(\bepsilon, \bphi)$ where $t(\cdot)$ is a differentiable function and $\bepsilon\sim p(\bepsilon)$ is a noise variable with fixed parameters.
With this parameterization, an unbiased differentiable minibatch-based Monte Carlo estimator of the expected data log-likelihood is obtained:
\small
\begin{equation}
	L_{\mc{D}}(\bphi)\simeq L_{\mc{D}}^{\mr{SGVB}}(\bphi)=\frac{N}{M}\sum_{i=1}^M\log p(\by_i|\bx_i,\bu=t(\bepsilon,\bphi)),
\end{equation}
\normalsize
where $\{(\bx_i, \by_i)\}_{i=1}^M$ is a minibatch of data with $M$ random instances $(\bx_i,\by_i)\sim \mc{D}$, and $\bepsilon \sim p(\bepsilon)$.
\subsection{Bayesian Nested Neural Network}
\label{sec:bn3}
\setlength{\belowdisplayskip}{2pt} \setlength{\belowdisplayshortskip}{2pt}
\setlength{\abovedisplayskip}{2pt} \setlength{\abovedisplayshortskip}{2pt}
In our model, the r.v. $\bu=(\bW,\bz)$ consists of two parts: weight matrix $\bW$ and ordering units $\bz$.
The ordering units order the network weights and generate sub-models that minimize the residual loss of a larger sub-model~\cite{rippel2014learning,ijcai2020-288}.
We define the corresponding variational parameters  $\bphi=(\btheta,\bbeta)$, 
where $\btheta$ and $\bbeta$ are the variational parameters for the weights and ordering units respectively.
%
We then have the following optimization objective,
\small
\begin{align}
	& \mc{L}_{\btheta,\bbeta}^{\mr{SGVB}}
	\simeq L_{\mc{D}}^{\mr{SGVB}}(\btheta,\bbeta)- \KL[q_{\btheta,\bbeta}(\bW,\bz)||p(\bW,\bz)],\label{eq:elbo2}
\\
\begin{split}
		 &L_{\mc{D}}^{\mr{SGVB}}(\btheta,\bbeta) =
		\\&\quad \frac{N}{M}\sum_{i=1}^M\log p(\by_i|\bx_i, \bW=t_w(\bepsilon_w,\btheta),\bz=t_z(\bepsilon_z,\bbeta)),\label{eq:lld}
		\end{split}
\end{align}
\normalsize
where 
$\bepsilon_z$ and $\bepsilon_w$ are the random noise, and 
$t_w(\cdot)$ and $t_z(\cdot)$ are the differentiable functions that transform the noises to the probabilistic weights and ordered masks.

Next, we focus on an example of a fully-connected (FC) layer.
Assume the FC 
layer in neural network takes in activations $\mathbf{H}\in\mathbb{R}^{M\times d}$ as the input, and outputs $\mathbf{F}=f(\mathbf{H})=\mathbf{H}\mathbf{W}$, where the weight matrix $\mathbf{W}\in\mathbb{R}^{d \times D}$, $d$ and $D$ are the input and output size, and $M$ is the batch size.
The elements are indexed as $h_{mi}$, $f_{mj}$ and $w_{ij}$ respectively.
We omit the bias for simplicity, and our formulation can easily be extended to include the bias term.
We have the ordering unit $\mb{z}\in\mbb{R}^{D}$ with each element $z_j$ applied on the column of $\mb{W}$, by which the columns of $\mb{W}$ are given different levels of importance.
Note that $\mb{z}$ is flexible, and can be applied to $\mb{W}$ row-wise or element-wise as well.

The prior for $\mb{W}$ assumes each weight is independent,  $p(\mb{W})=\prod_{ij}p(w_{ij})$, where $i\in\{1,\dots,d\}$ and $j\in\{1,\dots,D\}$.
We choose to place a mixture of two univariate variables as the prior over each element of the weight matrix $w_{ij}$.
For example, if we use the univariate normal distribution, 
then each $w_{ij}$ is a Gaussian mixture, where the 2 components are:
\begin{align}
p(w_{ij}|z_j=0)=\mc{N}(w_{ij}|\mu_{ij}^0,{\sigma_{ij}^0}^2)\nonumber\\
p(w_{ij}|z_j=1)=\mc{N}(w_{ij}|\mu_{ij}^1,{\sigma_{ij}^1}^2)\nonumber
\end{align}
where $(\mu_{ij}^0, \sigma_{ij}^0)$ and $(\mu_{ij}^1, \sigma_{ij}^1)$ are the means and standard deviations for the two components.
We fix $\mu_{ij}^0=0$ and $\sigma_{ij}^0$ to be a small value, resulting in a spike at zero for the component when $z_j=0$.
The variable $z_j$ follows the chain of Bernoulli distributions proposed in (\ref{eq:berchain}).
Using (\ref{eq:marginal_z}), the marginal distribution of $w_{ij}$ is then 
\begin{align}
    &p(w_{ij}) 
= \nonumber\\ &(1-\prod_{k=1}^i\pi_k) \mc{N}(w_{ij}|\mu_{ij}^0,{\sigma_{ij}^0}^2)
 + (\prod_{k=1}^i\pi_k) \mc{N}(w_{ij}|\mu_{ij}^1,{\sigma_{ij}^1}^2).\nonumber
\end{align}
%


To calculate the expected data log-likelihood, our \emph{Downhill} distribution allows efficient sampling and differentiable transformation for the ordering units (Section~\ref{sec:vou}).
The reparameterization of weight distributions has been widely studied~\cite{kingma2015variational, louizos2017multiplicative, kingma2013auto} to provide gradient estimate with low variance.
Our framework is compatible with these techniques, which will be discussed in Section~\ref{impl}.
An overview of sampling is shown in Figure~\ref{fig:overview}.
\subsection{Posterior Approximation}
\label{approx}
\setlength{\belowdisplayskip}{2pt} \setlength{\belowdisplayshortskip}{2pt}
\setlength{\abovedisplayskip}{2pt} \setlength{\abovedisplayshortskip}{2pt}
Next, we introduce the computation of the KL divergence.
We assume the posterior $q_{\btheta}(\mb{W})$ takes the same form as the prior, while $q_{\bbeta}(\mb{z})$ takes the $\mr{Downhill}$ distribution $\mb{z}\sim\mr{Downhill}(\bbeta,\tau)$.
We consider the case that $\tau\rightarrow 0$ for simplicity, while $\tau$ can be adjusted in the training process as annealing.
%
%
For this layer, the KL divergence in (\ref{eq:elbo2}) is
\begin{align}
& \KL[q_{\bbeta,\btheta}(\mb{W},\mb{z})||p(\mb{W},\mb{z})]\label{eq:reg}\\
&=\underbrace{\mathbb{E}_{q_{\bbeta}(\mathbf{z})}[\log\frac{q_{\bbeta}(\mathbf{z})}{p(\mathbf{z})}]}_{\Phi_1}
+\underbrace{\mathbb{E}_{q_{\bbeta}(\mb{z})}\mathbb{E}_{q_{\btheta}(\mb{W}|\mb{z})}[\log\frac{q_{\btheta}(\mb{W}|\mb{z})}{p(\mb{W}|\mb{z})}]}_{\Phi_2}.\nonumber
\end{align}
Term $\Phi_1$ of (\ref{eq:reg}) is
\begin{align}
\Phi_1  
=\sum_{\bz\in\calV} q_{\bbeta}(\mb{z})\log\frac{q_{\bbeta}(\mb{z})}{p(\mb{z})}=\sum_{j=1}^{D} \KL[q_{\bbeta}(\bv_j)||p(\bv_j)],\nonumber 
\end{align}
where $\calV = \{\bv_1,\dots,\bv_D\}$ is the set of ordered masks.
The number of components in the $\bz$ space is reduced from $D^2$ to $D$, because there are only $D$ possible ordered masks.
%
By definition, the probabilities are
\begin{align}
q_{\bbeta}(\bv_j) = \beta_j, \quad 
p(\bv_j) = (1-\pi_{j+1})\prod_{k=1}^j \pi_k,
\end{align}
where we define $\pi_{D+1}=0$.
Then we can write 
\begin{equation}
\Phi_1=\bbeta^T(\log(\bbeta)-\log((\mathbf{e}-\tilde{\boldsymbol{\pi}})^T\boldsymbol{J}_L\boldsymbol{\pi})),\label{eq:mat-phi1}
\end{equation}
where $\log(\cdot)$ is an element-wise log function, $\boldsymbol{\pi}=[\pi_1,\dots,\pi_D]$ and $\tilde{\boldsymbol{\pi}}=[\pi_2,\dots,\pi_{D+1}]$.


%

We define $\kappa_{ij}^k(\btheta)$ as the KL of $w_{ij}$ for component $\kappa\in \{0,1\}$.
Consider the matrices $\bs{\kappa}^0_{\btheta}=[\kappa_{ij}^0(\btheta)]_{ij}\in\mbb{R}^{d\times D}$ and $\bs{\kappa}^1_{\btheta}=[\kappa_{ij}^1(\btheta)]_{ij}\in\mbb{R}^{d \times D}$, which are easily computed by applying the KL function element-wise.
The term $\Phi_2$ is then expressed as
\begin{align}
\Phi_2 = 
	\mathbf{e}^T \bs{\kappa}^0_{\btheta} (\mb{J}-\mb{J}_L)^T \bs{\beta}+
	\mathbf{e}^T\bs{\kappa}^1_{\btheta} \mb{J}_L^T \bs{\beta},\label{eq:mlp-kl-2b}
\end{align}
where $\mathbf{e}$ is a vector of 1s, $\mb{J}$ is a matrix of 1s and $\mb{J}_L$ is a lower triangular matrix with each element being 1.

\textbf{Ordered $\ell_0$-Regularization.} We show that, if given the spike-and-slab priors, our KL term in (\ref{eq:reg}) has an interpretation as \emph{a generalization of an ordered $\ell_0$ regularization over the sub-networks}.
The corresponding reduced objective for deterministic networks is 
\begin{equation}
\min_{\btheta,\bbeta}\mbb{E}_{q(\bz|\bbeta)}L_\mc{D}(\btheta,\bbeta)+\lambda\sum_j^D j\beta_j \label{eq:ell0}
\end{equation}
Note that larger sub-networks have greater penalization.
The derivations and proofs for $p(\bv_j)$, $\Phi_2$ and regularization are in Appendix~\ref{ap-proof}.

\subsection{Implementation}
\label{impl}

\begin{figure}[t]
	\begin{center}
		\includegraphics[width=0.8\linewidth]{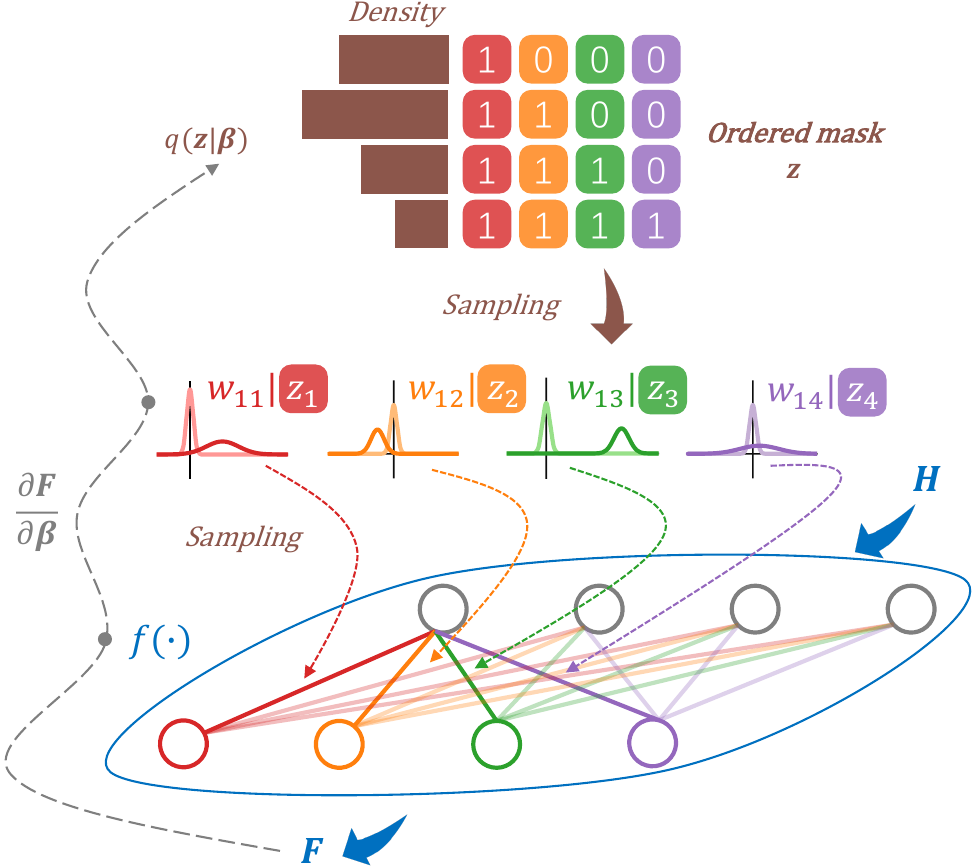}
	\end{center}
	\caption{Sampling process in a layer for calculating the data log-likelihood~(Equation~\ref{eq:lld}). A fully connected layer $f(\cdot)$ takes $\mb{H}$ as an input and outputs $\mb{F}$. The variational ordering unit $q(\bz|\bs{\beta})$ generates ordered mask $\bz=[z_j]_j$. Nodes $w_{ij}$'s with the same color share an element $z_j$. The gradient through stochastic nodes $\frac{\partial \mb{F}}{\partial \bs{\beta}}$ can be estimated efficiently, to update the importance $\bs{\beta}$.} 
	\label{fig:overview}
\end{figure}

For efficient sampling of the weight distributions, we put multiplicative Gaussian noise $\eta_{ij}\sim\mc{N}(1,\alpha)$ on the weight $w_{ij}$, similar to~\cite{kingma2015variational,molchanov2017variational,louizos2017bayesian}. We take $w_{ij}$ for $z_j=1$ as an example.
\begin{align}
	w_{ij}&=\theta_{ij}\eta_{ij} =\theta_{ij}(1+\sqrt{\alpha_{ij}}\epsilon_w), \ \epsilon_w\sim\mc{N}(0,1), \\
	w_{ij}& \sim\mc{N}(w_{ij}|\theta_{ij},\alpha_{ij}\theta_{ij}^2).
\end{align}
%
We also assume a log-uniform prior~\cite{kingma2015variational,molchanov2017variational,louizos2017bayesian}, i.e., $p(\log |w_{ij}|\,\mid z_j=1)=\mr{const}$.
With this prior, 
the negative KL term $-K_{ij}(\btheta)^1$ does not depend on the  variational parameter $\theta_{ij}^1$~\cite{kingma2015variational}, when the parameter $\alpha_{ij}$ is fixed,
%
%
\begin{align}
	&-\KL[q(w_{ij}|\theta_{ij}^1,\alpha_{ij}, z_j=1)||p(|w_{ij}|\, | z_j=1)]\nonumber\\
	&\quad\quad =\frac{1}{2}\log\alpha_{ij}-\mbb{E}_{\epsilon_w\sim\mc{N}(1,\alpha_{ij})}\log |\epsilon_w|+C, \label{eq:alpha_func}
\end{align}
where $C$ is a constant.
Note that the prior can be flexibly replaced by other distributions like Gaussian, while we choose log-uniform for simplicity of optimization, as $\theta_{ij}$ is eliminated from the computation of KL.

As the second term in (\ref{eq:alpha_func}) cannot be computed analytically and should be estimated by sampling, Kingma~\emph{et al}~\cite{kingma2015variational} propose to sample first and design a function to approximate it.
But their approximation of $K_{ij}^1(\theta)$ does not encourage $\alpha_{ij}>1$ as the optimization would be difficult. 
They truncate $\alpha_{ij}\leq 1$ corresponding to a small variance, which is not flexible.
Molchanov~\emph{et al}~\cite{molchanov2017variational} use a different parameterization that pushes $\alpha_{ij}\rightarrow\infty$, as illustrated in Figure~\ref{fig:alpha}.
This means the $w_{ij}$ can be pruned, generating a single sparse neural network.
In our model, we want the order or sparsity of weights to be explicitly controlled by the ordering unit $\mb{z}$, otherwise the network would collapse to a single model rather than generate a nested set of sub-models.
Thus, we propose another approximation to $-K_{ij}(\btheta)^1$ (\ref{eq:alpha_func}),
\begin{align}
	a_1 e^{-e^{a_4}\cdot(a_2+a_3*\log\alpha_{ij})^2}-0.5\log(1+\alpha_{ij}^{-1}) +C, \label{eq:our_alpha}
\end{align}
where $a_1=0.7294$, $a_2=-0.2041$, $a_3=0.3492$ and $a_4=0.5387$.
We obtained these parameters by sampling from $\bepsilon_w$ to estimate (\ref{eq:alpha_func}) as the ground-truth and fit these parameters for $10^5$ epochs.
For fitting the curves, the input range is limited to $\log\alpha\in[-5,0.5]$.
As shown in Figure~\ref{fig:alpha}, our parameterization allows $\alpha>1$ and maximizing $-\KL$ does not push $\alpha$ to infinity (c.f. 
\cite{kingma2015variational} and 
\cite{molchanov2017variational}), providing \emph{more flexible} choices for the weight variance.

\begin{figure}[t]
	\begin{center}
		\includegraphics[width=0.7\linewidth]{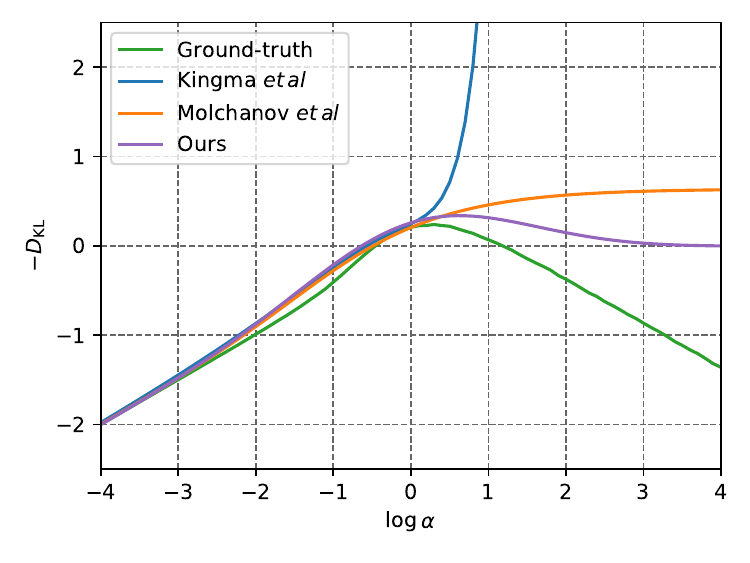}
	\end{center}
	\caption{Approximation to (\ref{eq:alpha_func}). Our approximation allows $\alpha>1$ (c.f., \cite{kingma2015variational}) and does not push $\alpha\rightarrow0$ to generate a collapsed model (c.f., \cite{molchanov2017variational}).}
	\label{fig:alpha}
\end{figure}

The pseudocde for training the proposed network is shown in Algorithm~\ref{algo:tr}.
For testing, only lines 2-7 are executed.

\setlength{\textfloatsep}{13pt}

\begin{algorithm}[t!]
	\scriptsize
	\caption{\footnotesize Pseudocode for training \BNNN.}
	\label{algo:tr}
	\KwIn{Parameters $\{\theta_{ij}^{(l)},\alpha_{ij}^{(l)},\beta_{j}^{(l)}\}_{ij}^{(l)}$, data $\{\mathbf{X}, \mathbf{Y}\}$, layer input $\mathbf{H}^{(0)}=\mathbf{X}_{\mr{batch}}$, total KL $\bar{\mr{KL}}=0$.}
	\begin{algorithmic}[1]
		\WHILE{the network is not converged}
		\FOR{$l=1:L$}
		\STATE Sample $\mathbf{b}^{(l)}=[b_{mj}^{(l)}]_{mj}$ with $\alpha_{ij}^{(l)}$, and (\ref{sampling-fc}) for dense layers or\\ (\ref{sampling-conv}) for convolution layers.
		\STATE Sample $\bz^{(l)}\sim q_{\bbeta^{(l)}}(\bz)$ and $\mathbf{F}^{(l)}=[f_{mj}^{(l)}]_{mj}=b_{mj}^{(l)}z_j^{(l)}$.
		\STATE Compute $\mr{KL}$ using (\ref{eq:reg2}), $\bar{\mr{KL}} \leftarrow \bar{\mr{KL}} +\mr{KL}$.
		\STATE $\mathbf{H}^{(l+1)}=\mathbf{F}^{(l)}$.
		\ENDFOR
		\STATE Compute loss as $\mathcal{L}_{\mc{D}}(\mathbf{Y}_{\mr{batch}},\mathbf{F}^{(l)})$, $\mathcal{L}=\mathcal{L}_{\mc{D}}-\bar{\mr{KL}}$.
		\STATE Compute $\frac{\partial\mathcal{L}}{\partial\theta}$, $\frac{\partial\mathcal{L}}{\partial\beta}$, $\frac{\partial\mathcal{L}}{\partial\alpha}$, update the network.
		\ENDWHILE
	\end{algorithmic}
\end{algorithm}

As the prior of the zero-component $w_{ij}$ is assumed a spike at zero with a small constant variance, we let $q(w_{ij}|z_j=0)$ be the same spike as in Section~\ref{sec:bn3} to save computation.
Also, to speed up the sampling process in Figure~\ref{fig:overview}, we directly multiply the sampled masks with the output features of the layer.
This saves the cost for sampling from $w_{ij}|z_j=0$ and simplifies (\ref{eq:mlp-kl-2b}) to $\mathbf{e}^T\bs{\kappa}^1_{\btheta} \mb{J}_L^T \bs{\beta}$.
Therefore, the KL divergence (\ref{eq:reg}) is simplified to 
\begin{align}
& \KL[q_{\bbeta,\btheta}(\mb{W},\mb{z})||p(\mb{W},\mb{z})]=\Phi_1+\Phi_2\label{eq:reg2}\\
&=\bbeta^T(\log(\bbeta)-\log((\mathbf{e}-\tilde{\boldsymbol{\pi}})^T\boldsymbol{J}_L\boldsymbol{\pi}))+\mathbf{e}^T\bs{\kappa}^1_{\btheta} \mb{J}_L^T \bs{\beta},\nonumber
\end{align}
where $\bs{\kappa}_{\btheta}^1$ is calculated by applying (\ref{eq:our_alpha}) element-wise.

Using the notation in Section~\ref{sec:bn3}, the output of a fully connected layer is 
\begin{align}
	f_{mj}=b_{mj}z_j, \quad b_{mj}\sim\mc{N}(\gamma_{mj}, \delta_{mj}), \label{sampling-fc}\\
	\gamma_{mj}=\sum_{i=1}^{d}h_{mi}\theta_{ij}, \quad \delta_{mj}=\sum_{i=1}^{d}h_{mi}^2\alpha_{ij}\theta_{ij}^2.\nonumber
\end{align}
The sampling process is similar to that of \cite{kingma2015variational,molchanov2017variational,louizos2017bayesian}.
This can be easily extended to convolutional layers with the ordering applied to channels (see Appendix~\ref{ap-impl-conv}).

\subsection{Distillation to Deterministic Nested Nets}

\begin{figure}[t]
	\begin{center}
		\includegraphics[width=0.8\linewidth]{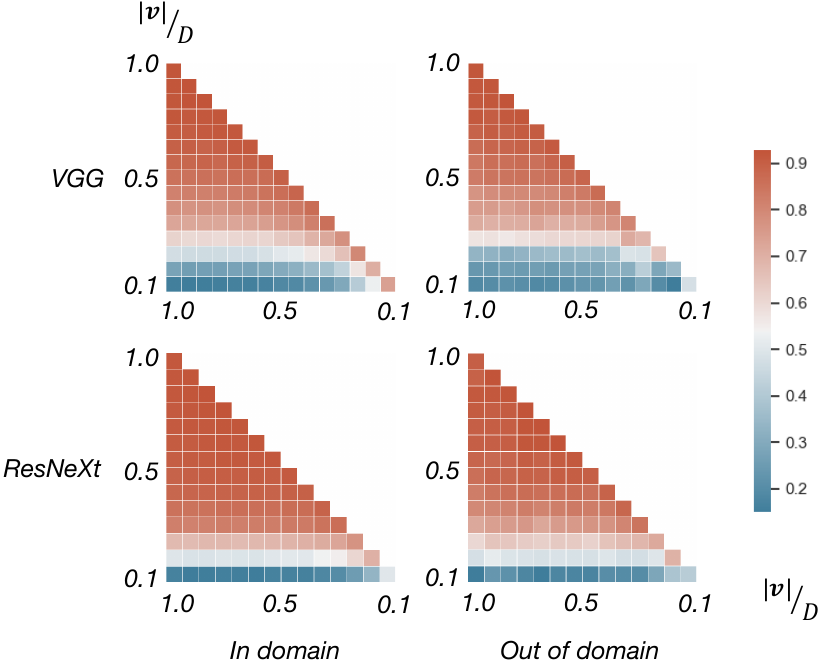}
	\end{center}
	\caption{The correlation between two output sample sets of \BNNN. VGG11 and ResNeXt are used as the backbones. The models are trained on the ``in domain'' data. The correlation are evaluated on both ``in domain'' and ``out of domain'' data. Red color indicates a high correlation.}
	\label{fig:distill_map}
\end{figure}

Using VND and the probabilistic treatment, the \BNNN is able to learn the importance of parameters from data.
With a decreasing order, the importance of parameters decays with an optimal trajectory instead of a hand-crafted one.
For pursuing a lighter neural network and a more practical inference speed, we can distill the Bayesian knowledge from \BNNN to its deterministic counterpart.
This is motivated by an analysis of correlation of outputs of \BNNN with different widths.
As shown in Figure~\ref{fig:distill_map}, we conduct correlation analysis on two \BNNN\, with VGG11 and ResNeXt as backbones.
For each network and a particular dataset, we obtain the uncertainty of the output under different widths, 
\begin{align}
&\mathcal{U}=[U_{\btheta|\bv}]_{\frac{|\bv|}{D}=0.1, 0.2,\dots,1.0}, \nonumber\\
&U_{\btheta|\bv}=[U_{\btheta|\bv,i}]_{i=1}^N=[U[\mathbb{E}_{q(\btheta|\bz)}[p(y|\bx_i,\btheta)]]]_{i=1}^N, \nonumber
\end{align}
where the $\frac{|\bv|}{D}=0.1$ means 10\% of \BNNN parameters are used.
We use $|\bv|$ to indicate the number of non-zero entries in $\bv$.
The conversion from $\bv$ to $\bz$ is illustrated in Section~\ref{approx}.
A common choice for $U$ is the entropy.
For calculating the correlation between the uncertainties [$U_{\btheta|\bv_i},U_{\btheta|\bv_j}$] for any pair $[\bv_i, \bv_j]$, the Pearson correlation coefficient is applied and visualized in Figure~\ref{fig:distill_map}.

\begin{figure}[t]
	\begin{center}
		\includegraphics[width=0.96\linewidth]{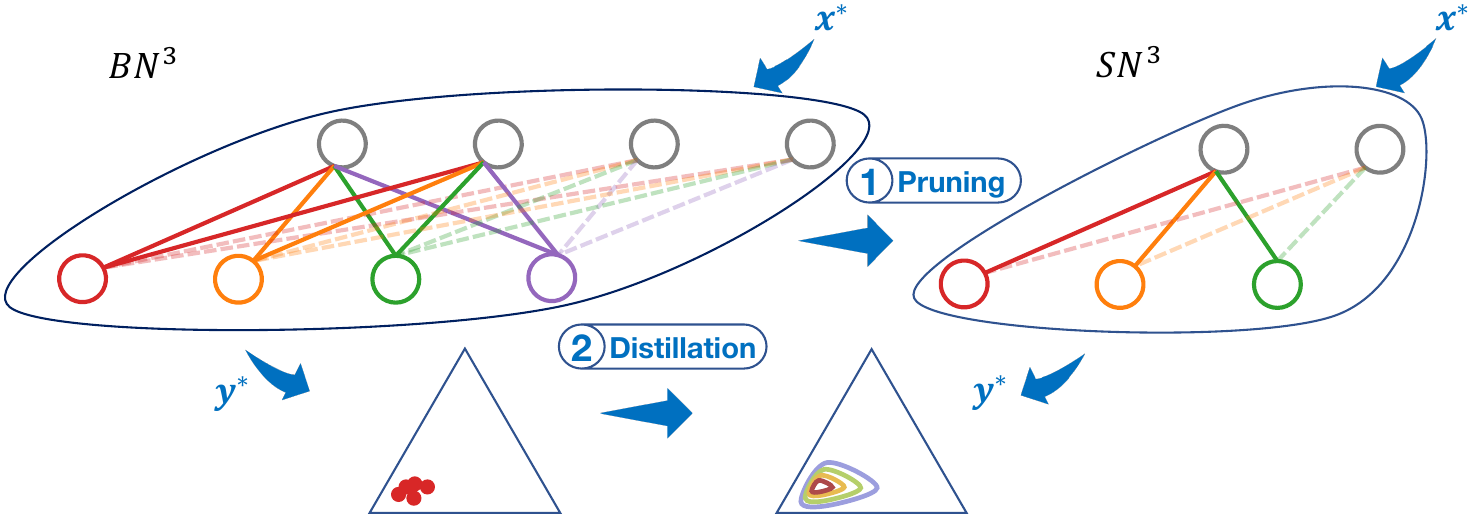}
	\end{center}
	\caption{The two step slimming to obtain a high performance deterministic network, $\mr{SN}^3$. The triangle represents a 2-simplex with 3 classes. The left red points are the output of \BNNN given input $\bx^\ast$, $\mr{BN}^3({\bx^\ast})$. The right contours represent the Dirichlet distribution parameterized by $\mr{SN}^3(\bx^\ast)$.}
	\label{fig:distill}
\end{figure}

With in-domain data, strong correlation of uncertainty could be observed with width greater than $0.2\sim0.25$.
This number is reduced to less than $0.2$ with the out-of-domain data.
This motivates us to prune the \BNNN, then distill the original \BNNN uncertainty to the pruned \BNNN for smaller model size and better performance, as shown in Figure~\ref{fig:distill}.
As a knowledge distillation process is adopted, we name it Student Nested Neural Network ($\mr{SN}^3$).

The first step is to prune the less useful weights according to the correlation analysis, to generate the base of $\mr{SN}^3$.
The mean of the nodes and the VND parameters from \BNNN are kept, while the variances in the variational parameters are discarded.

The second step produces a distillation from a Bayesian neural network to a deterministic neural network.
For better capturing the uncertainty from \BNNN, we adapt the loss function introduced in~\cite{cui2020accelerating} to our problem, which distills the dark knowledge from a Bayesian neural network (\BNNN) to a parameterized Dirichlet ($\mr{SN}^3$).
Specifically, a Maximum Mean Discrepancy (MMD) is used as the probabilistic distance metric, between the generated samples from \BNNN and a Dirichlet distribution parameterized by the output of $\mr{SN}^3$.
\begin{align}
    &\mr{MMD}_\mathcal{K}(q,\mr{Dir})\\
    =&\sup_{\substack{\Psi\in\mc{H}_{\mc{K}},\\||\Psi||_{\mc{H}\leq 1}}}\mbb{E}_{q(\bar{\bz})}[\mbb{E}_{q(\theta|\bz)}[\Psi(\mr{BN}^3(\bx))]-\mbb{E}_{s\sim \mr{Dir}({\mr{SN}^3}_{\bar{\bz}}(\bx))}[\Psi(s)]]\nonumber
\end{align}
Note that $\mc{H}_{\mc{K}}$ is a reproducing kernel Hilbert space defined by a positive-definite kernel $\mc{K}$.
$\Psi$ is known as the critic.
$\mr{Dir}$ is the Dirichlet distribution whose parameters are given by the output of $\mr{SN}^3$.

Different from the formulation in \cite{cui2020accelerating}, during the distillation, the size of $\mr{SN}^3$ and \BNNN should vary together, controlled by sampling from $q(\bar{\bz})$ ($q(\bar{\bv})$), where 
$\bar{\bz}$ and $\bar{\bv}$ are the VND random variable for the pruned $\mr{SN}^3$.
$\bv$ is no longer sampled from $q(\bv)$ but determined by $\bar{\bv}$
\begin{equation}
|\bv| = r|\bar{\bv}|, 
\end{equation}
\abc{where $|\bv|$ is the number of dimensions in vector $\bv$, }
and $r$ is the ratio of parameters between \BNNN and $\mr{SN}^3$.
For example, suppose a layer in \BNNN has 16 nodes, the corresponding layer of $\mr{SN}^3$ is initialized with 4 nodes by the correlation analysis, then $r=16/4=4$.
%

%

%

\section{VND for Generative Models}
\label{sec:vnd_generative}
Variational nested dropout encodes the importance of dimensions and optimizes by learning from data.
Another interesting branch is to order the latent distributions of generative models.
The latent variables with specified orders are expected to implicitly learn information with different levels of importance.
Due to the stochastic nature of the variational nested dropout, we study two Bayesian deep generative models: variational auto-encoder and probabilistic UNet.

\subsection{Ordered Encoding Variational Bayes}

\subsubsection{Overview}
\label{sec:vnd_ov}
The proposed variational nested dropout is useful for organizing the latent space of the generative models, like variational auto-encoders.
We propose VND-AE, which generates diverse samples compared by organizing the latent space.

To introduce the methodology, we show the basic form of prior  using a \emph{Gaussian-Bernoulli chain} distribution as the latent distribution for a generative model.
\yedit{The prior is based on the $\rm{BernChain}(\bz,\bpi)$, introduced in Section~\ref{sec:bou}, with random variable $\bz=[z_1,\dots,z_D]$ for the mask and $\bpi=[\pi_1,\dots,\pi_D]$ for the parameters.
%
}
We let a uni-variate Gaussian variable conditioned on each node of the Bernoulli chain,
\begin{align}
    p(h_i \,| \,z_i) = z_i\,\, \mc{N}(h_i | 0, \sigma^2) + \,\,(1 - z_i)\,\, \mc{N}(h_i |0,\sigma_\infty^2) \label{eq:mix}
\end{align}
where $z_i$ is sampled from the above Bernoulli chain.
%
$h_i$ is the $i$-th latent representation for $\bx$.
$\sigma_\infty^2$ represents a constant variance with large value.
Note that, the log-uniform prior (Section~\ref{impl}) or other distributions could also adopted for the conditional variable $h_i|z_i$ here.

\begin{figure}[t]
	\begin{center}
		\includegraphics[width=0.5\linewidth]{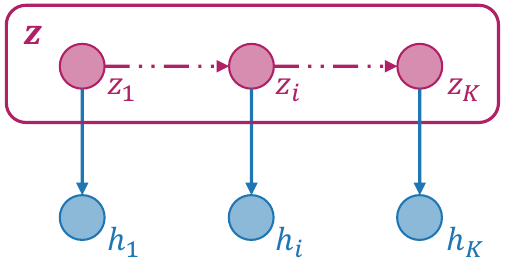}
	\end{center}
	\caption{The graphical representation of the Gaussian-Bernoulli chain prior.}
	\label{fig:alpha}
\end{figure}

Each node of the latent representation could be regarded as a spike-and-slab model~\cite{murphy2012machine}.
Traditionally, \abc{such models use \emph{independent} Bernoulli r.v.s on the latent $z_i$'s to control the sparsity through randomly ``selecting'' an $h_i$.} 
\abc{However, due to the independence,} there is no specific order for the selected variables.
Different from the traditional models, our proposed prior is chained by the sequence of Bernoulli variables.
As a result, there is a strict order due to the Bernoulli chain, as the $(i-1)$-th failure would stop the sampling of subsequent variables.

The approximate posterior are univariate Gaussian variables conditioned on the Downhill distribution presented in Section~\ref{sec:vou}.
The sampling process of the posterior is
\begin{align}
\bz|\bx\sim\rm{Downhill}(\cdot|f^\bz(\bx),\tau) \nonumber\\
h_i|z_i, \bx \sim z_i\,\, \mc{N}( h_i | f_i^{\bmu} (\bx), f_i^{\bsg^2} (\bx)) + (1-z_i)\,\, \mc{N}( h_i | 0, \sigma_\infty^2)
\end{align}
where $f_i^{\bmu} (\cdot)$ and $f_i^{\bsg} (\cdot)$ are the encoder network for mean and variance of latent variables, \yedit{while $f^\bz(\bx)$ generates parameters for the Downhill distribution.}
%
%

As the Downhill distribution is fully re-parameterizable, we  let $\bz$ be conditioned on $\bx$ for learning a flexible representation.
The latent variables (codes) for different inputs may have varying effective latent dimension (code length).
Thus, the VND-VAE is trained to learn the optimal code length for different input $\bx$.
The evidence lower bound (ELBO) is written as:
\begin{align}
\log p(\bx)\geq \underbrace{\mathbb{E}_{\bh,\bz}[\log p(\bx|\bh, \bz)]}_{\rm{reconstruction \,\, term}}-\underbrace{\rm{KL}(q(\bh,\bz|\bx)||p(\bh,\bz))}_{\rm{KL\,\,term}}
\end{align}

\begin{figure}[t]
	\begin{center}
		\includegraphics[width=0.9\linewidth]{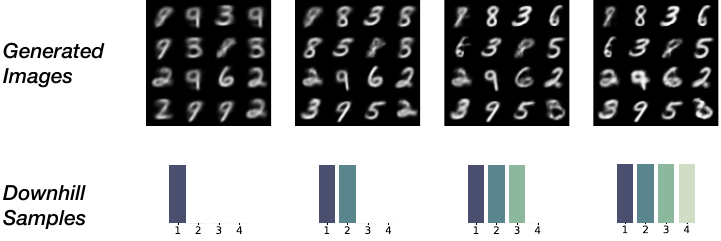}
	\end{center}
	\caption{\abc{Digit images generated from 16 samples of $\bz$ (the 4x4 grid of digits) using different latent variable (code) lengths (increasing from left to right).}}
	\label{fig:mnist_diverse}
\end{figure}

A toy example on the image generation task is shown in Figure~\ref{fig:mnist_diverse}, \abc{where different length codes are used for generation.}
During testing time, we use fixed Downhill samples to control the code length, by which the effective dimension of $\bh$ is controlled.
We can observe, the first dimensions generate blurred ``meta digits'' with regular fonts and similar styles.
With longer code length, the digits are sharpened and transformed to different styles.
Thus, the VND-AE realizes the encoding of ordered information into the latent probabilistic distributions.

We present detailed discussion of the KL divergence in Appendix~\ref{appx:elbo}, to understand how the VND affects the training objective of variational auto-encoder.

\subsubsection{Discussion of Diversity}
\label{sec:divers}
For the single-modal prior like Gaussian with diagonal matrix as the covariance matrix, the diversity is poor as the posterior will seek to approximate to that mode even if a positive lower bound is given (Appendix~\ref{appx:elbo}).
Our prior has a mixture structure as shown in Figure~\ref{fig:mnist_diverse} and (\ref{eq:mix}).
Different from a vanilla mixture, the first several modes are weighted with higher importance, which decays with increasing index value.

The posterior shares similar multi-modal property with the prior, but with higher efficiency for sampling.
During training, the approximation is done by selecting a chain of modes with a strict order under the mode-seeking reverse KL.
When the posterior is multi-modal, the diversity is achieved.
A simple guarantee is that, when the optimal code length is greater than 1 (the case of \emph{single-modal}), the diversity could be achieved.
The \emph{single-modal} case corresponds to $\beta_1=1$ and $\beta_i=0, \forall i>1$.
We show this case \emph{never} happens due to our design of posterior in Appendix~\ref{appx:elbo}.
The approximation that collapses to a single mode is thus avoided.

\begin{figure}[t]
	\begin{center}
		\includegraphics[width=0.98\linewidth]{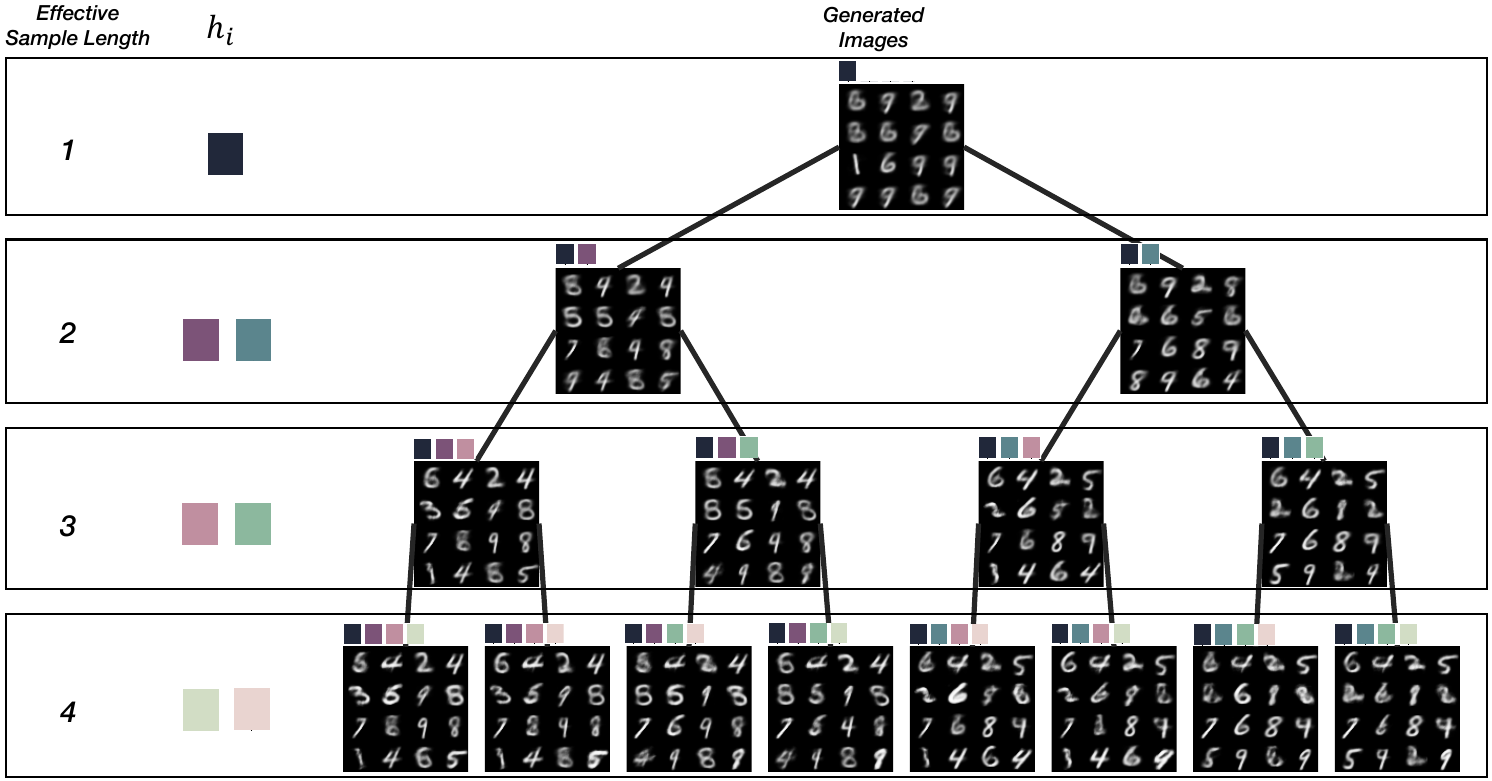}
	\end{center}
	\caption{The generated images given specific samples of $\bh$. Different colored blocks for $h_i$ represent different values sampled from a univariate Gaussian. The group of colored blocks at the top left corner of a image represents the sample $\bh$. In this example, the maximum length of $\bh$ is 4.}
	\label{fig:sample_tree}
\end{figure}

The diversity is empirically illustrated via the toy example used in Section~\ref{sec:vnd_ov}, with a different sample generation method, as shown in Figure~\ref{fig:sample_tree}.
After training VND-AE, the samples $\bh$ are drawn from the latent space with fixed Downhill samples $\bz$.
Except for the first dimension, we give each dimension $i$ two choices of $h_i$.
This forms a tree structured data generation process, which conforms to the \abc{induced nested (hierarchical)} structure of the latent space.
%
At each level, with the increase of the sample length, the generated images are sharpened and refined.
More importantly, different $h_i$'s at level $i$ leads to different writing styles or output results.
The data generation process in this toy example also conforms with the training process; as the first several levels with shorter code lengths are sampled more often, \abc{the corresponding outputs in the first levels are more generic.}
VND-AE progressively enriches the diversity by controlling the code length.
Samples at a greater code length level are more diverse, as it involves more modes in the mixture.
By comparion, VAE and its variants see a single-modal multivariate distribution during the whole training process, which is not beneficial for the diversity.

\subsection{Ordered Encoding Aleatoric Uncertainty}
\label{sec:vndpunet}
The probabistic UNet~\cite{kohl2018probabilistic} and its variants have shown the unique property of encoding the aleatoric uncertainty.
The methodology is to jointly use a deterministic neural network and a VAE whose prior is conditioned on the input $\bx$.
The posterior is conditioned on both $\bx$ and $\by$ such that it could help capture the noise from data.
This is useful when there are observable noise from the labels, e.g., the case when multiple experts provide labels with disagreement for the same input.

We apply VND to the probabilistic UNet to effectively  bring ordered information to its latent distributions.
%
We name this variant VND-PUNet where ``P'' refers to probabilistic.
As the prior is not a fixed probability distribution and requires gradients, we let it be a parameterized Downhill distribution in our VND-PUNet,
%

\begin{align}
\label{eq:punet_pr1}
\bz|\bx&\sim \abc{p(\bz|\bx) =} \rm{Downhill}(f^{\bz,\rm{pr}}(\bx),\tau),  \\
h_i\abc{|z_i,\bx} &\sim p(h_i|z_i,\bx) = z_i\, \mc{N}( h_i | f_i^{\bmu,\rm{pr}} (\bx), f_i^{\bsg^2,\rm{pr}} (\bx)), \label{eq:punet_pr}
\end{align}
\abc{where $f^{\bz,\rm{pr}}, f_i^{\bmu,\rm{pr}}, f_i^{\bsg^2,\rm{pr}}$ are the encoder networks for the prior.}
By taking $\by$ as the input for the VAE part, the posterior is written as,
\begin{align}
\bz|\bx,\by&\sim \abc{q(\bz|\bx,\by) =} \rm{Downhill}(f^{\bz,\rm{po}}(\bx,\by),\tau), \\
h_i\abc{|z_i,\bx,\by} &\sim q(h_i|z_i,\bx,\by) \nonumber \\ &\quad=z_i\, \mc{N}( h_i | f_i^{\bmu,\rm{po}} (\bx,\by), f_i^{\bsg^2,\rm{po}} (\bx,\by)),
\end{align}
\abc{where $f^{\bz,\rm{po}},f_i^{\bmu,\rm{po}}, f_i^{\bsg^2,\rm{po}}$ are the encoder networks for the posterior.}

\yedit{We adapt the training objective in~\cite{kohl2018probabilistic} to the VND-PUNet}, 
\begin{align}
\begin{split}
\abc{{\cal L}(\bx,\by)} &=
\abc{\mathbb{E}_{\bz\sim q(\bz|\by,\bx)}[
\mathbb{E}_{\bh\sim q(\bh|\bz,\by,\bx)}}[\log p(\by|f_{\mr{base}}(\bx,\bh)]] \\
&\quad -\rm{KL}(q(\bh,\bz|\by,\bx)||p(\bh,\bz|\bx))
\end{split}
\end{align}
where $f_{\mr{base}}$ is the deterministic neural network, e.g., a UNet, which takes $\bx$ and the samples from the latent space as the input.
The KL terms is written as
\begin{align}
\begin{split}
& \KL[q(\bh,\bz|\by,\bx)||p(\bh,\bz|\bx)] \\
&=\sum_i f_i^{\bz,\rm{po}}(\bx,\by)\Big[\log\frac{f_i^{\bz,\rm{po}}(\bx,\by)}{f_i^{\bz,\rm{pr}}(\bx)},
+\kappa_i(\bx,\by) \Big]
\end{split}
\end{align}
where $\kappa_i(\bx,\by)$ is the KL divergence between two univariate Gaussian variables of $h_i$.
During testing time, the samples are obtained from the prior distribution via \abc{(\ref{eq:punet_pr1}-\ref{eq:punet_pr}),
$\bz^\ast\sim p(\bz|\bx^\ast),\bh^\ast\sim p(\bh|\bz^\ast,\bx^\ast)$}, and  then fed to the deterministic neural network to obtain the results $p(\by^\ast|f_{\mr{base}}(\bx^\ast,\bh^\ast))$.

\section{Related work}
\label{sec:related}
In this section, we reviewed  deep nets with $\ell_0$-regularization and nested nets, while the comparisons with Bayesian neural network are elaborated in Section~\ref{impl}.
The related generative models are discussed, as well.

\textbf{$\ell_0$-regularization.} The Bernoulli-Gaussian linear model with independent Bernoulli variables is shown to be equivalent to $\ell_0$ regularization~\cite{murphy2012machine}.
Recent works~\cite{louizos2018learning,yang2019deephoyer} investigate $\ell_0$ norm for regularizing DNNs.
\cite{louizos2018learning} presents a general formulation of a $\ell_0$-regularized learning objective for a single deterministic neural network, 
\begin{equation}
\min_{\tilde{\btheta},\tilde{\bs{\pi}}}\mbb{E}_{q(\tilde{\bz}|\tilde{\bs{\pi}})}[{L_D(\tilde{\btheta})}]+\lambda\sum_{j=1}^{|\tilde{\btheta}|}\tilde{\pi}_j,
\end{equation}
where the variable $\tilde{\bz}$ is a binary gate with parameter $\tilde{\pi}_j$ for each network node $\tilde{\theta}_j$, and $L_D$ is the loss.
It was shown that $\ell_0$-regularization over the weights is 
a special case of an ELBO over parameters with spike-and-slab priors.
These works present the \emph{uniform} $\ell_0$-regularization as the coefficient $\lambda$ is a constant over the weights.
It is interesting that our training objective (\ref{eq:elbo2}) can be viewed as a generalization of a new training objective of deterministic networks, 
%
which includes a weighted penalization over the choices of sub-networks, interpretable as \emph{an ordered $\ell_0$-regularization} (\ref{eq:ell0}).


\textbf{Nested neural networks.}
Nested nets have been explored in recent years, for its portability in DNN deployment on different platforms.
%
\cite{kim2018nestednet} proposes a network-in-network structure for a nested net. which consists of internal networks from the core level to the full level.
%
\cite{yu2018slimmable,yu2019universally} propose slimmable NN that trains a network that samples multiple sub-networks of different channel numbers (widths) simultaneously, where the weights are shared among different widths.
The network needs to switch between different batch normalization parameters that correspond to different widths.
To alleviate the interference in optimizing channels in slimmable NN, 
\cite{cai2019once} proposes a once-for-all network that is elastic in kernel size, network depth and width, by shrinking the network progressively during training.
%
\cite{ijcai2020-288} proposes using \emph{nested dropout} to train a fully nested neural network, which generates more sub-networks in nodes, including weights, channels, paths, and layers.
However, \emph{none} of the previous works consider learned importance over the nodes and the predictive uncertainty.
Our work provides a well-calibrated uncertainty and the learned importance, with a full Bayesian treatment of nested nets.

\textbf{Variational auto-encoder.}
The variational auto-encoder (VAE) was introduced in~\cite{kingma2013auto} to bring scalability and training stability for the variational learning of latent variable models.
The reparameterization trick for the latent variables and stochastic variational Bayes framework provides the fundamental support for the scalable variational inference.
$\beta$-VAE~\cite{higgins2016beta} augments the VAE with a single hyper-parameter $\beta$ which helps to quantify the degree of learnt disentanglement.
Joint-VAE~\cite{dupont2018learning} further augments a discrete latent unit for discovering the categorical factors from the latent space.
Other works study different probabilistic distance metrics like Wasserstein distance~\cite{tolstikhin2018wasserstein}.
Sliced-Wasserstein auto-encoder~\cite{kolouri2018sliced} regularizes the auto-encoder loss with the sliced-Wasserstein distance between the distribution of the encoded training samples and a samplable prior distribution.
Different from these works, our VND-AE builds an explicit ordered structure over the latent distributions, showing better diversity while maintaining a high generation performance.

\textbf{Probabilistic UNet.}
The family of probabilistic UNet~\cite{kohl2018probabilistic} is designed for capturing the aleatoric uncertainty, when the labels of data are noisy.
For example, there are multiple doctors provides segmentation masks for the same histopathological image.
The VAE part of probabilistic UNet encodes both the input and output into the posterior.
The epistemic uncertainty is captured by pairing a UNet as the backbone for segmentation.
During testing time, only the prior is used for sampling different possible segmentation masks.
Our VND-PUNet place the order structure into the latent space of the input image and labels, which helps organize the underlining latent space that learns aleatoric uncertainty.

\section{Experiments}
\label{sec:exp}
We next present experiments \abc{showing that VND can be applied to a variety of model architectures and tasks.}
The experiments include three main tasks: image classification using \BNNN and $\mr{SN}^3$, image generation for VND-AE, and probabilistic semantic segmentation for VND-PUNet.
%


\subsection{Image Classification with \BNNN}

\begin{figure*}
\centering
\includegraphics[width=0.7\linewidth]{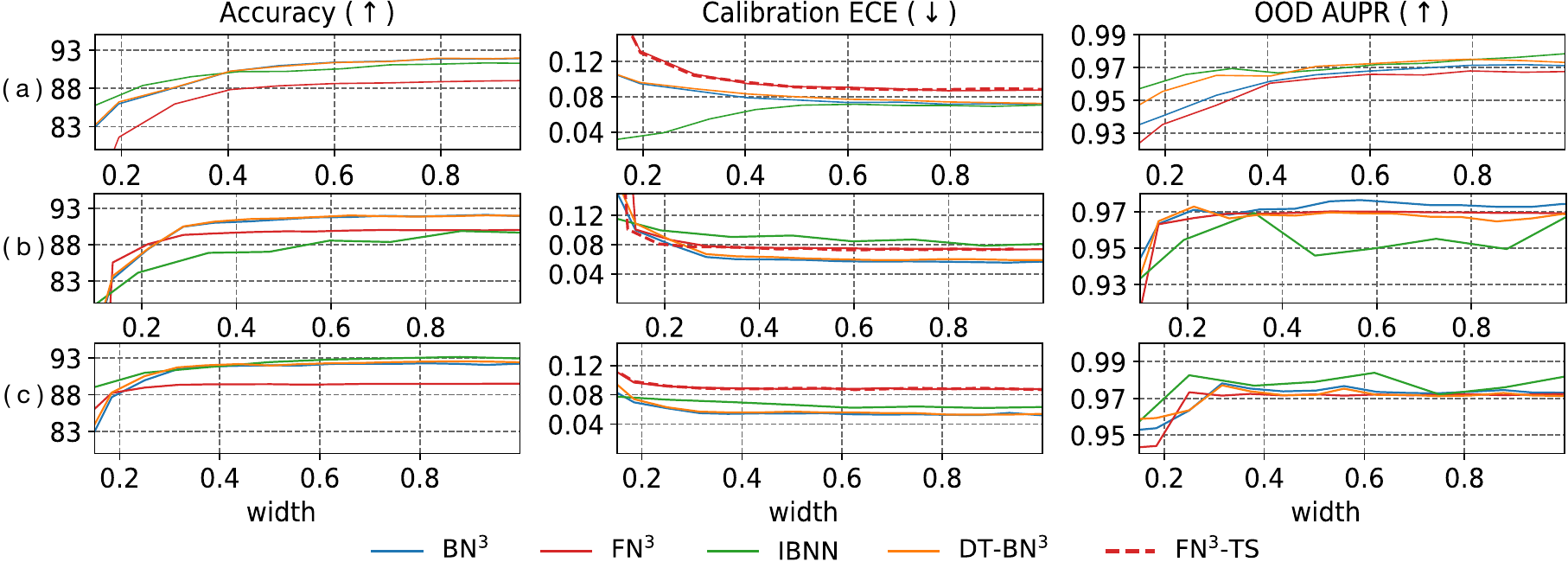}
\caption{Results on Cifar10 for (a) VGG11, (b) MobileNetv2, and (c) ResNeXt-Cifar. Each curve plots performance versus the network width.
}
\label{fig:cifar10}
\end{figure*}
\begin{figure*}
\centering
\includegraphics[width=0.7\linewidth]{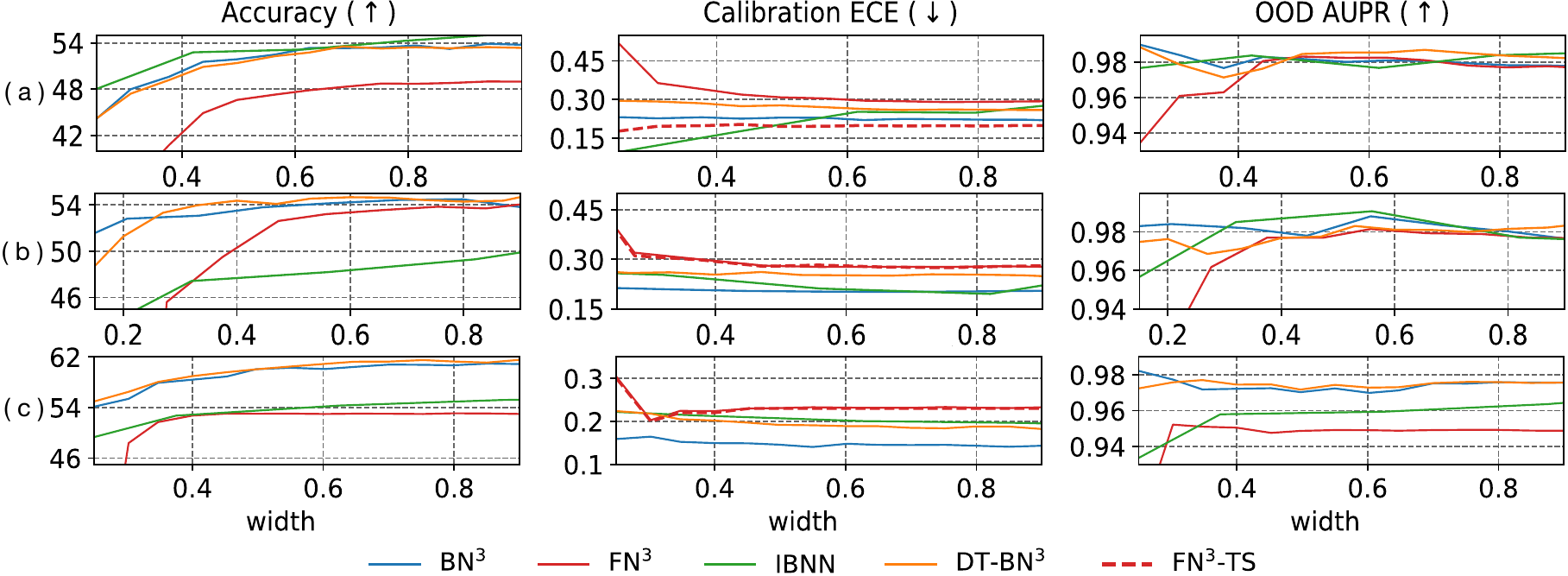}
\caption{Results on Tiny ImageNet for (a) VGG11, (b) MobileNetv2,  (c) ResNeXt-Cifar.
}
\label{fig:tinynet}
\end{figure*}

\subsubsection{Experiment setup}

\textbf{Dataset and Models.} The image classification experiments are conducted on Cifar10 
SVHN and Tiny Imagenet (see Appendix~\ref{ap-exp-cifar10} for results on Cifar100).
%
The tested NN models are VGG11 with batch normalization layers~\cite{simonyan2014very}, ResNeXt-Cifar model from~\cite{xie2017aggregated}, and MobileNetV2~\cite{sandler2018mobilenetv2}.

\textbf{Methods.} To train the proposed Bayesian nested neural network (denoted as \BNNN), we use the cross-entropy loss for the expected log-likelihood in (\ref{eq:lld}).
The computation of the KL term follows Section~\ref{impl}.
For ordering the nodes, in every layer, we assign each dimension of the prior (Bernoulli chain) and posterior (Downhill variable) of the ordering unit to a group of weights.
Thus, the layer width is controlled by the ordering unit.
We set the number of groups to 32 for VGG11 and ResNext-Cifar, and to 16 for MobileNetV2.
We compare our $\mr{BN}^3$ with the fully nested neural network ($\mr{FN}^3$)~\cite{ijcai2020-288}, since it is an extension of slimmable NN~\cite{yu2018slimmable,yu2019universally} to fine-grained nodes.
We also compare with the Bayesian NN with variational Gaussian dropout~\cite{kingma2015variational}, where
 we train a set of independent Bayesian NNs (IBNN) 
for different fixed widths.
Conceptually, the performance of IBNN, which trains separate sub-networks, is the ideal target for \BNNN, which uses nested sub-networks.
%
In addition, the deterministic \BNNN\,(DT-\BNNN) which only uses the mean of posterior is considered.
For comparing the calibration performance, we conduct the temperature scaling~\cite{guo2017calibration} using the sub-networks of \FNNN (\FNNN-TS).

During testing time, we generate fixed width masks for $\mr{BN}^3$ and $\mr{FN}^3$ as in~\cite{ijcai2020-288}.
We re-scale the node output by the probability that a node is kept (see Appendix~\ref{ap-exp-rescaling}).
The batch normalization statistics are then re-collected for 1 epoch with the training data (using fewer data is also feasible as shown in Appendix~\ref{ap-exp-bn}).
The number of samples used in testing \BNNN and IBNN is 6.
%
The detailed hyper-parameter settings for training and testing are in Appendix~\ref{ap-exp-setup}.

\textbf{Evaluation metrics.}
For the evaluation, we test accuracy, uncertainty calibration, and out-of-domain (OOD) detection.
Calibration performance is measured with the expected calibration error~\cite{guo2017calibration} (ECE), which is the expected difference between the average confidence and accuracy.
\yedit{The OOD detection performance is evaluated on the SVHN dataset.}
OOD performance is measured with the area under the precision-recall curve (AUPR) \cite{boyd2013area, hendrycks2016baseline, lakshminarayanan2017simple} (see Appendix~\ref{ap-exp-ood} for AUROC curves).
If we take the OOD class as positive, precision is the fraction of detected OOD data that are true OOD, while recall is the fraction of true OOD data that are successfully detected.
%
Note that a better model will have higher accuracy and OOD AUPR, and lower calibration ECE.
%
As the sampling and collection of batch-norm statistics are stochastic,
we repeat each trial 3 times and report the average results.

\subsubsection{Results}
The results are presented in Figs.~\ref{fig:cifar10} and \ref{fig:tinynet}.
%
First, looking at performance versus width, $\mr{BN}^3$ exhibits the well-behaved property of sub-networks, where the performance increases (accuracy and AUPR increase, ECE decreases) or is stable as the width increases. This demonstrates that the variational ordering unit successfully orders the information within each layer.

Despite learning nested sub-networks, in general, $\mr{BN}^3$ has similar performance as IBNN (which separately learns sub-networks) for all models and datasets, with the following exceptions.
For MobileNetV2 on both datasets, $\mr{BN}^3$ outperforms IBNN in all metrics, as IBNN fails to perform well in prediction and uncertainty (outperformed by $\mr{FN}^3$ too).
%
%
%
For VGG11 on both datasets, IBNN tends to have lower ECE with smaller widths, showing its advantage in providing uncertainty for small and simple models.
However, IBNN has larger ECE when the model size is large, e.g., 
$\mr{BN}^3$ has lower ECE than IBNN with the ResNeXt model.
%
%
%
Finally, 
$\mr{BN}^3$ outperforms IBNN by a large margin for ResNeXt on Tiny ImageNet, 
which we attribute 
to its ability to prune the complex architecture via learning ordered structures (Section~\ref{sec:vou}) and the ordered $\ell$-0 regularization effect (Section~\ref{approx}),  which are absent in IBNN.
Comparing the two nested models,  $\mr{BN}^3$ outperforms $\mr{FN}^3$ in all metrics, which shows the advantage of learning the nested dropout rate for each node.
The temperature scaling method (\FNNN-TS) improves calibration under few settings, but is not scalable when the number of sub-networks is large.

\textbf{Efficiency.} 
We compare \yedit{the performance with limited number of parameters} of \BNNN\,with the sparse variational Bayes (SVB)~\cite{molchanov2017variational} on Cifar10 with VGG11 in Table~\ref{tab:efficiency}.
\BNNN\,with smallest width outperforms SVB, and has a similar size.
MC inference (\BNNN-MC) averages the predictions from the stochastic forward passes of different widths, and performs slightly worse than full-width \BNNN \yedit{but uses fewer parameters}.

\begin{table}[h!]
    \caption{Performance with limited parameters, SVB, Monte-Carlo integration and full-width \BNNN.}
	\label{tab:efficiency}
	\begin{center}
		\begin{small}
			\centering
			\begin{tabular*}{0.47\textwidth}{@{}c@{}|c|c|c|@{}c@{}}
			&Acc.($\uparrow$)&ECE($\downarrow$)&OOD($\uparrow$)& Weight usage (\%)\\
				\hline
				smallest \BNNN& 83.5 & 0.10 & 0.94 & 6.2\\
				SVB~\cite{molchanov2017variational}& 82.9  & 0.17 & 0.94 & 5.7\\
				\hline
				\BNNN-MC & 89.4 & 0.08 & 0.97 & {46\textsubscript{mean} $\pm$ 27\textsubscript{std} } \\
				\BNNN & 92.6 & 0.07& 0.97& 100
			\end{tabular*}
		\end{small}
	\end{center}
\end{table}

Training of \BNNN\,is twice as slow as training a {\em single} BNN, but much faster than training IBNNs for large number of widths $N$. BNN test inference is slow when using sampling, but can be sped up by using the parameters' mean (denoted as DT-\BNNN) with similar performance (see Figures~\ref{fig:cifar10} and \ref{fig:tinynet}).

\begin{table}[h!]
\caption{Training and testing time for \BNNN/DT-\BNNN, IBNN and \FNNN.}
	\label{time}
	\begin{center}
		\begin{small}
			\begin{tabular}{c | c| c | c}
				&\BNNN/ DT-\BNNN&IBNN&\FNNN\\
				\hline
				Train& 14.9 h (\BNNN) & 7.5$\times$\,$N$ h & 5.6 h\\
				Test& 1.78 s (DT-\BNNN) & 1.68 s & 1.82 s\\
			\end{tabular}
		\end{small}
	\end{center}
\end{table}

\noindent\textbf{Knowledge distillation.}
We perform the knowledge distillation experiments on the harder Tiny Imagenet dataset.
We test two teacher networks, VGG11 and ResNeXt-Cifar, as their number of channels is power of 2, which is suitable for pairing the teacher network and student network.
According to the correlation analysis, we keep 25\% of channels for $\mr{SN}^3$ for both networks.
For uncertainty calibration, we leave the last layer of $\mr{SN}^3$ to be a Bayesian layer with stochastic nodes described in Section~\ref{impl}.
The numerical results are shown in the following table.

\begin{table}[h!]
\caption{Performance of $\mr{SN}^3$ with different sizes and relative sizes.}
	\begin{center}
			\begin{tabular}{c | c | c@{\hspace{0.1cm}}c@{\hspace{0.1cm}}c}
			Model& Rel. Size & Accuracy(\%)&ECE&OOD AUPR\\
				\hline
            VGG11-\BNNN-100\%& 1& 54.00 & 0.22 & 0.980 \\
			VGG11-$\mr{SN}^3$-25\%&1/16& 54.98 & 0.21 & 0.981 \\
			VGG11-$\mr{SN}^3$-50\%&1/8& 55.26 & 0.21 & 0.981 \\
			VGG11-$\mr{SN}^3$-75\%& 3/16& 55.32 & 0.20 & 0.980 \\
			VGG11-$\mr{SN}^3$-100\%&1/4& \textbf{55.36} & \textbf{0.20} & \textbf{0.982} \\
			\midrule
            ResNeXt-\BNNN-100\% & 1 & 61.19 & 0.15 & 0.977 \\
			ResNeXt-$\mr{SN}^3$-25\% &1/16 & 63.10 & 0.15 & 0.977 \\
			ResNeXt-$\mr{SN}^3$-50\% & 1/8& 63.52 & 0.15 & 0.976 \\
			ResNeXt-$\mr{SN}^3$-75\% & 3/16& 63.78 & 0.14 & \textbf{0.980} \\
			ResNeXt-$\mr{SN}^3$-100\% & 1/4& \textbf{63.90} & \textbf{0.14} & \textbf{0.980} \\
			\end{tabular}
	\end{center}
\end{table}
Note that the size of $\mr{SN}^3$-100\% is $1/4$ of \BNNN-100\%.
With the correlation analysis, model pruning and distillation method, the $\mr{SN}^3$ \abc{with the smallest width (1/16 of the original size)} outperforms \BNNN.
%
By setting the last layer to be a stochastic layer with our Bayesian treatment, the uncertainty calibration and out-of-domain detection performance are improved through knowledge distillation.

\begin{figure*}[t]
\centering
    \includegraphics[width=\linewidth]{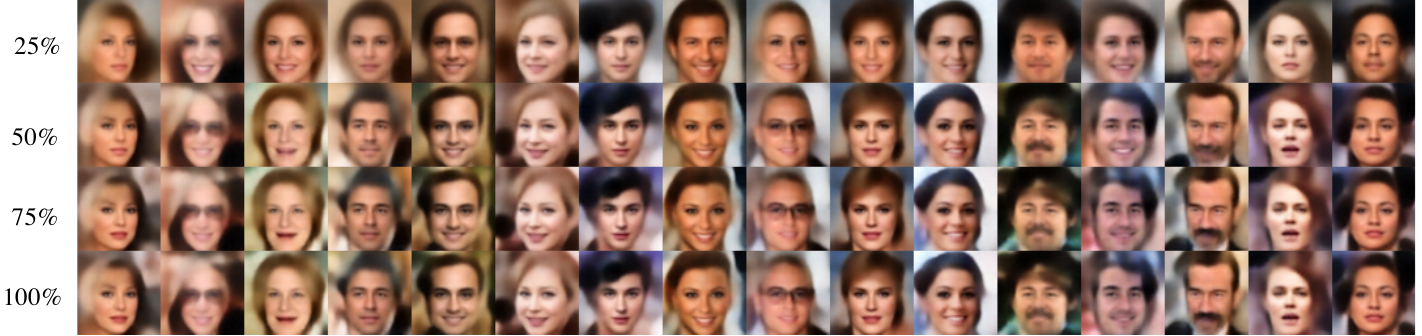}
    \caption{Generated faces by VND-AE for 25\%, 50\%, 75\%, 100\% latent dimension (code length). \abc{As the code length increases, the face images become more detailed.}
    }
    \label{fig:faces}
\end{figure*}

\begin{figure}[htbp]
\centering
    \includegraphics[width=\linewidth]{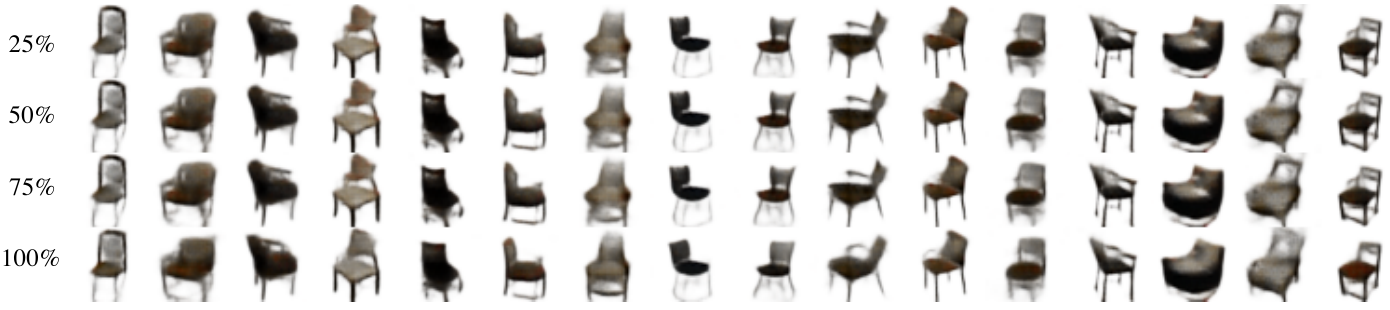}
    \caption{VND generated chairs for 25\%, 50\%, 75\%, 100\% sample length from top to bottom. Each row shows generated chair with corresponding sample length.}
    \label{fig:chairs}
\end{figure}

\begin{figure*}[h!]
\centering
    \includegraphics[width=0.8\linewidth]{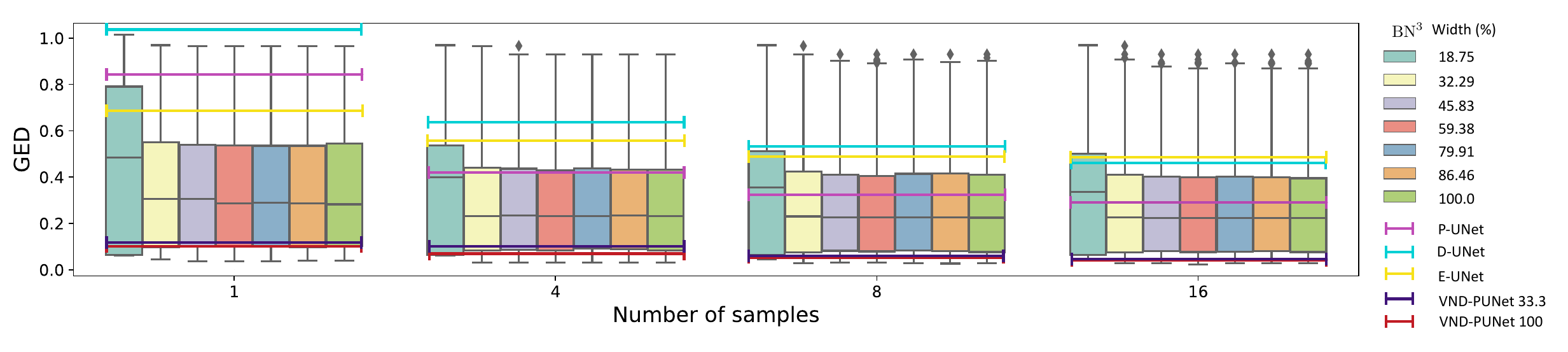}
    \caption{Evaluation on semantic segmentation using generalized energy distance ($\downarrow$) with different numbers of posterior samples.
    Each box-plot shows the GED of all data of $\mr{BN}^3$ for one network width (\%).  The black horizontal line in the box plot represents the mean.  The bold horizontal lines represent the averaged results for comparison methods (using full width) and VND-PUNet \abc{(using 33\% or 100\% width).}}
    \label{fig:segm}
\end{figure*}
\subsection{Image Generation with VND-AE}

\subsubsection{Experiment setup}
\textbf{Dataset.} 
The image generation experiments are conducted on five datasets across image domains: CelebA~\cite{liu2015faceattributes}, Cifar10~\cite{krizhevsky2009learning}, Cifar100~\cite{krizhevsky2009learning}, 3D Chairs~\cite{Aubry14} and Chest X-ray~\cite{ChestImages}. CelebA is a large-scale face dataset with more than 200K celebrity images. 3D Chairs contains a total of 86,366 synthesized images sampled from 1,393 high-quality 3D chair models with 62 different viewpoints. Chest X-ray dataset contains 5,856 validated Chest X-Ray images.

\textbf{Evaluation Metrics.} 
We adopt the \emph{Frechet Inception Distance score} (FID)~\cite{heusel2017gans}, \emph{Inception Score} (IS)~\cite{salimans2016improved} and \emph{Reconstrction Error} as the evaluation metrics. 
FID measures the distance between feature vectors extracted by Inception Network for real and generated images.
IS evaluates the quality of synthetic images output by generative models by measuring the KL divergence of generated images.

\textbf{Methods}
We use a standard Variational Auto-Encoder (VAE) with five convolutional layers following with BatchNorm as the VAE backbone, and proposed VND-AE is developed based on this structure as well. 
LeakyRelu is selected as the activation function, and the full size latent dimension (code length) is set as 128.
We compare a group of four latent dimensions [32, 64, 96, 128] of VND-AE and VAE, which corresponding to 25\%, 50\%, 75\% and 100\% of full latent dimension size, respectively.
The training of VND-AE uses a model with latent dimension 128, and in testing time, we only truncate the latent dimension to the specified length. 
%
The training of VAE requires re-initializing the model and training for each particular latent dimension.
Comparison with other VAE variants (BetaVAE~\cite{higgins2016beta}, JointVAE~\cite{dupont2018learning} and SWAE~\cite{kolouri2018sliced}) are also presented, and 
their latent dimension is also 128.

\begin{table}[tbp]
  \centering
  \caption{Frechet Inception Distance (FID) scores for image generation. Lower FID is better.}
  \setlength{\tabcolsep}{1.3mm}{
    \begin{tabular}{cc|c|c|c|c|c}
    Model & Dim.      & \multicolumn{1}{l|}{CelebA} & \multicolumn{1}{l|}{Cifar10} & \multicolumn{1}{l|}{Cifar100} & \multicolumn{1}{l|}{3D Chairs} & \multicolumn{1}{l}{Chest X-ray} \\
    \hline
    BetaVAE&100\% & 77.33 & 215.04 & 154.44 & 183.67 & 316.02 \\
    JointVAE& 100\% & 67.95 & 154.57 & 154.67 & 134.31 & 296.57 \\
    SWAE& 100\% & 78.33 & 119.52 & 118.26 & 109.15 & 369.72 \\
    \hline
    \multirow{4}[2]{*}{VAE} & 25\%  & 72.29 & 175.17 & 162.46 & 170.93 & 300.36 \\
          & 50\%  & 72.43 & 140.70 & 161.71 & 178.84 & \textbf{282.62} \\
          & 75\%  & 73.40 & 144.35 & 142.15 & 173.89 & 300.49 \\
          & 100\% & 73.14 & 147.35 & 146.25 & 175.94 & 298.53 \\
    \hline
    \multirow{4}[1]{*}{VND} & 25\%  & 76.54 & 174.83 & 154.97 & 94.87 & 287.07 \\
          & 50\%  & 64.09 & 134.95 & 123.71 & 96.57 & 285.79 \\
          & 75\%  & 62.29 & 117.24 & 113.00 & 96.48 & 285.79 \\
          & 100\% & \textbf{62.80} & \textbf{114.12} & \textbf{111.95} & \textbf{96.44} & 285.73 \\
    \end{tabular}%
    }
  \label{tab:fid}%
\end{table}%

\subsubsection{Results}

Tables \ref{tab:fid}, \ref{tab:is}, \ref{tab:recon_err} present the results on the 5 datasets for the 3 metrics.
In general, VND-AE outperforms the VAE baseline for most metrics, and it requires \emph{training only once} to obtain models with different latent dimensions.
By comparisons, the VAE baseline requires repetitively training for different latent dimensions.
Under most metrics and datasets, VND-AE with 50\% latent dimensions already outperforms BetaVAE, JointVAE and SWAE.

\textbf{Frechet Inception Distance score (FID).}
A lower FID indicates a closer distance between the pile of generated and real images.
We can conclude from Table~\ref{tab:fid} that increasing latent dimension leads to better FID for both VAE and VND-AE, and VND-AE is generally superior. 
Although constructed with the same network structure, the ordered latent variables generated by VND-AE can achieve smaller FID than VAE. 
One exception is on the Chest X-ray dataset, VAE with 50\% latent dimension outperforms VND-AE, while other latent dimensions have a large gap with VND-AE.
Note that VND-AE provides consistently decent performance under different latent dimensions for this dataset.
The images generated by VND-AE has better overlap with the distribution of the original image set. 
When latent dimensions are the same, VND-AE's FID score outperforms the other baselines across five datasets from various domains.

\begin{table}[tbp]
  \centering
  \caption{Inception scores (IS) for image generation. Higher IS is better.}
  \setlength{\tabcolsep}{1.3mm}{
    \begin{tabular}{cc|c|c|c|c|c}
    Model & Dim.      & \multicolumn{1}{l|}{CelebA} & \multicolumn{1}{l|}{Cifar10} & \multicolumn{1}{l|}{Cifar100} & \multicolumn{1}{l|}{3D Chairs} & \multicolumn{1}{l}{Chest X-ray} \\
    \hline
    BetaVAE&100\% & 1.60  & 2.19  & 2.90  & 3.37  & 1.08 \\
    JointVAE& 100\% & \textbf{1.85}  & 3.01  & 2.83  & 3.47  & 1.08 \\
    SWAE& 100\% & 1.70  & 3.00  & 2.74  & 3.24  & 1.20 \\
    \hline
    \multirow{4}[2]{*}{VAE} & 25\%  & 1.68  & 2.71  & 2.67  & 3.07  & 1.11 \\
          & 50\%  & 1.68  & 3.26  & 2.60  & 3.05  & 1.12 \\
          & 75\%  & 1.66  & 3.19  & 3.19  & 3.04  & 1.12 \\
          & 100\% & 1.65  & 3.10  & 2.99  & 3.04  & 1.02 \\
    \hline
    \multirow{4}[1]{*}{VND} & 25\%  & 1.56  & 2.83  & 2.73  & 3.51  & 1.28 \\
          & 50\%  & 1.69  & 3.36  & \textbf{3.32}  & \textbf{3.54}  & 1.28 \\
          & 75\%  & 1.74  & \textbf{3.40}  & 3.23  & \textbf{3.54}  & 1.27 \\
          & 100\% & 1.75  & 3.36  & 3.15  & 3.53  & \textbf{1.29} \\
    \end{tabular}%
    }
  \label{tab:is}%
\end{table}%

\textbf{Inception score (IS).} Higher IS indicates better diversity of the generated images. 
A comparison of VND-AE and VAE of different latent dimensions and other baselines is shown in  Table~\ref{tab:is}. 
Like the observation on FID, a larger latent dimension leads to better IS performance, which means a greater diversity of images generated by the model. 
VND-AE outperforms VAE on this metric on each dimension, although VND-AE is trained only once while VAE had to be re-trained for each  latent dimension. 
VND-AE outperforms other methods with higher IS on all datasets, except for JointVAE on CelebA.
\yedit{This might be due to JointVAE has the advantage of using a mixture of discrete codes and continuous codes for diversity.}
Extending VND-AE to discrete codes could be interesting future work.

\begin{table}[tbp]
  \centering
  \caption{Reconstruction Error for image generation.  Lower error is better.}
    \setlength{\tabcolsep}{1.3mm}{
    \begin{tabular}{c|c|c|c|c|c|c}
    Model & Dim.      & \multicolumn{1}{l|}{CelebA} & \multicolumn{1}{l|}{Cifar10} & \multicolumn{1}{l|}{Cifar100} & \multicolumn{1}{l|}{3D Chairs} & \multicolumn{1}{l}{Chest X-ray} \\
    \hline
    BetaVAE&100\% & 1.338 & 1.286 & 0.842 & 0.904 & 1.823 \\
    JointVAE&100\% & 1.320 & 0.774 & 0.882 & 0.753 & 1.833 \\
    SWAE&100\% & 4.264 & 3.820 & 3.882 & 1.426 & 10.58 \\
    \hline
    \multirow{4}[2]{*}{VAE} & 25\%  & 1.277 & 0.947 & 1.048 & 1.360 & 1.517 \\
          & 50\%  & 1.276 & 0.752 & 0.984 & 1.382 & 1.561 \\
          & 75\%  & 1.277 & \textbf{0.740} & 0.913 & 1.381 & 1.523 \\
          & 100\% & 1.279 & 0.743 & 0.827 & 1.380 & 1.529 \\
    \hline
    \multirow{4}[1]{*}{VND} & 25\%  & 1.299 & 1.094 & 1.131 & 0.661 & 0.856 \\
          & 50\%  & 1.052 & 0.865 & 0.899 & 0.618 & 0.836 \\
          & 75\%  & 1.008 & 0.810 & 0.843 & 0.618 & 0.832 \\
          & 100\% & \textbf{1.004} & 0.805 & \textbf{0.838} & \textbf{0.618} & \textbf{0.832} \\
    \end{tabular}%
    }
  \label{tab:recon_err}%
\end{table}%

\textbf{Qualitative study.} Besides the improvement on standard generation metrics, VND-AE also offers the possibility of component-by-part generation.
Figure~\ref{fig:faces} shows sample face images generated by proposed VND-AE, for 4 latent dimensions (code lengths). 
The first row corresponds to 25\% code length, and the increasing the code length (the next rows) introduces additional details to the previously generated image. 
These details include, but are not limited to, facial makeup, sunglasses, beard, mouth opening, gender switching, etc. 
This function is caused by VND-AEs's explicit order constraints on the latent dimension during training. 
By adding order to the latent dimension during training, \abc{each group of dimensions learns specific features from common to infrequent.}
%
The component-by-part generation gives a way to produce images of a person with different expressions, angles and accessories in just one training process, or add artificial details to an already existing synthesized face.
This property is also evident when jointly examining the reconstruction errors in Table~\ref{tab:recon_err} and IS.
Generated images with 25\% latent dimensions are with relatively low IS (low diversity), showing a ``meta face'' which is possible to be refined by adding more details.
\abc{These ``meta faces'' at 25\% also have high reconstruction error compared to the original images, since the original faces are detailed.} 
%
As another example, Figure~\ref{fig:chairs} shows how VND controls chair generation. 
Similar to face generation, VND adds more details when increasing the code length, including filling the back of the chair, adding chair legs, etc.

\subsection{Lung Abnormalities Segmentation with VND-PUNet}
\subsubsection{Experiment setup}
\textbf{Dataset and Evaluation Metrics.} The semantic segmentation experiments are conducted on the LIDC-IDRI~\cite{clark2013cancer} dataset, which 
contains 1,018 CT scans from 1,010 lung patients with manual lesion segments annotated by four experts.
This dataset is uncertainty-critical as it contains typical ambiguities in labels that appear in medical applications.
We follow \cite{kohl2018probabilistic} to process the data, resulting in 12,870 images in total.
We adopt the \emph{generalized energy distance} (GED)~\cite{bellemare2017cramer,salimans2018improving,szekely2013energy,kohl2018probabilistic} as the evaluation metric, with $\delta(A,B)=1-\mr{IoU}(A,B)$ as the distance function \abc{between two segmentation map distributions $A, B$}.  GED measures the distance between the output distributions rather than single deterministic predictions. 
For \BNNN, it measure the probabilistic distances between the induced distribution from model posterior given a fixed width, and the noisy labels from the four experts.

\textbf{Models.} We use a standard UNet~\cite{ronneberger2015u} for $\mr{BN}^3$ and the number of groups is 32.
We also present the results of VND-PUNet proposed in  Section~\ref{sec:vndpunet}.
We compare with Probabilistic UNet (P-UNet)~\cite{kohl2018probabilistic}, a deep ensemble of UNet (E-UNet), and Dropout UNet (D-UNet)~\cite{kendall2015bayesian}. Their results are the average results from~\cite{kohl2018probabilistic} with the full UNet.


\subsubsection{Results}
The results are presented in Figure~\ref{fig:segm}.
We observe that $\mr{BN}^3$ outperforms the existing methods in most of the cases, with the difference more obvious when there are fewer posterior samples.
%
The performance of $\mr{BN}^3$ stabilizes after width of 32.29\%.
This indicates $\mr{BN}^3$ learns a compact and effective structure compared with other methods, in terms of capturing ambiguities in the labels.
%


When there are more posterior samples (8 and 16), probabilistic UNet has better performance than the $\mr{BN}^3$ with  the smallest width ($6/32$ channels are preserved).
This means with more posterior samples, the probabilistic UNet can depict the latent structure better, but requires a full-width model.
Increasing the width to 32.29\%, $\mr{BN}^3$ then achieves better performance.

The VND-PUNet consistently outperforms other variants of probabilistic UNet and \BNNN with either $1/3$ or 100\% latent dimensions.
The diversity of generated samples are effectively enhanced by adopting the VND, \abc{resulting in better modeling of uncertainty.}

\section{Conclusion}
\label{sec:conclusion}
In this paper, we propose a novel variational nested dropout (VND) that explicitly models the ordered information via our proposed Downhill random variable.
%
From our model, the ordered information can be learned from data, rather than hand-tuned as with previous methods \abc{like nested dropout}.
We validate the effectiveness of the proposed VND on two applciations: constructing nested nets and variational generative models.
Experiments show that VND can improve both accuracy and calibrated predictive uncertainty for the nested nets.
It can enhance the performance of VAE in image reconstruction and diversity.
\yedit{On the tasks of capturing annotation noise by encoding aleatoric uncertainty, VND-PUNet and \BNNN outperform several variants of UNet and PUNet.}
Future work will study the VND in language modeling and sequential data where the ordering is also important, e.g., \cite{rai2015large}.
The Downhill random variable is a well-suited hidden variable for such applications.
%




{\small
\bibliographystyle{ieee_fullname}
\bibliography{egbib}
}

\newpage

\onecolumn
\title{Variational Nested Dropout (Appendix)}

\maketitle


\renewcommand{\thesection}{\Alph{section}}
\renewcommand{\theequation}{A\arabic{equation}}
\setcounter{equation}{0}
\setcounter{section}{0}

\section{Derivation and Proofs}
\label{ap-proof}
\subsection{Derivation of Properties}
\setcounter{figure}{7}
\label{ap-proof-property}

\textbf{Property 1.} If $\mathbf{c}\sim \mathrm{Gumbel\_softmax}(\tau,\beta,\epsilon_z)$\footnote{For Gumbel-softmax sampling, we first draw $g_1 \dots g_K$ from $\mathrm{Gumbel}(0,1)$, then calculate $c_i=\mr{softmax}(\frac{\log(\beta_i)+g_i}{\tau})$. The samples of $\mr{Gumbel}(0,1)$ can be obtained by first drawing $\epsilon_z\sim \mr{Uniform}(0,1)$ then computing $g=-\log(-\log(\epsilon_z))$.}, then $z_i=1-\mathrm{cumsum}_i^\prime(\mathbf{c})$, where $\mathbf{e}$ is a $K$-dimensional vector of ones, and $\mathrm{cumsum}_i^\prime(\mathbf{c})=\sum_{j=0}^{i-1}c_j$. $c_0\coloneqq1$. $\epsilon_z$ is a standard uniform variable.

We show that using the sampling process in Property 1 recovers produce the Downhill random variable.
We assume $\mathbf{c}$ follows a Gumbel softmax distribution~\cite{gumbel1948statistical, maddison2014sampling} which has the following form.
\begin{align}
p(c_1,\dots,c_K)
=\Gamma(K)\tau^{K-1}(\sum_{i=1}^K \pi_i/c^\tau)^{-K}\prod_{i=1}^K(\pi_i/c^{\tau+1})
\end{align}
We apply the transformation $T_i(\cdot)=\mathbf{e}_i-\mathrm{cumsum}_i^\prime(\mathbf{\cdot})$ to the variable $\mathbf{c}$.
$\mathbf{z}=T(\mathbf{c})=\mathbf{e}-\mathrm{cumsum}_i^\prime(\mathbf{c})$

To obtain the distribution of $p(\mathbf{z})$, we apply the change of variables formula on $\mathbf{c}$.
\begin{align}
p(\mathbf{z})&=p(T^{-1}(\mathbf{z}))\Big|\mathrm{det}(\partial \frac{T^{-1}(\mathbf{z})}{\partial \mathbf{z}})\Big|\\
p(z_{1:K})&=p(T^{-1}(z_{1:K}))\Big|\mathrm{det}( \frac{\partial T^{-1}(z_{1:K})}{\partial z_{1:K}})\Big|
\end{align}
From the definition of $T(\cdot)$, we can obtain $T_i^{-1}(\mathbf{z})=z_{i-1}-z_i$, and its 
Jacobian is
\begin{align}
\frac{\partial T^{-1}(z_{1:K})}{\partial z_{1:K}}=
\begin{bmatrix}
-1 & 0  & \dots & 0 & 0 \\
1  & -1 & \dots & 0 & 0 \\
0  & 1  & \dots & 0 & 0 \\
\vdots & & & & \vdots \\
0  & 0  & \dots & 1 & -1
\end{bmatrix}
\end{align}
Thus, $|\mathrm{det}(\frac{\partial T^{-1}(z_{1:K})}{\partial z_{1:K}})|=1$.
Finally, we have
\begin{align}
p(z_{1:K})=p(T_{1:K}^{-1}(\mathbf{z}))
=\Gamma(K)\tau^{K-1}(\sum_{i=1}^K\frac{ \pi_i}{(z_{i-1}-z_i)^\tau})^{-K}\prod_{i=1}^K(\frac{\pi_i}{(z_{i-1}-z_i)^{\tau+1}}).
\end{align}

\textbf{Property 2.} When $\tau\rightarrow 0$, sampling from the Downhill distribution reduces to discrete sampling, where the sample space is the set of ordered mask vectors $\calV$.
The approximation of the Downhill distribution to the Bernoulli chain can be calculated in closed-form.

As shown in~\cite{maddison2016concrete,jang2016categorical}, 
when $\tau\rightarrow 0$, the Gumbel softmax transformation corresponds to an argmax operation that generates a one-hot vector:
\begin{equation}
\mb{c} = \mr{one\_hot}(\mr{argmax}_i(g_i+\log\beta_i)),
\end{equation}
where the relative order is preserved.

Say a sample $\mb{c}^\ast\sim \mr{Gumbel\_softmax}(\bbeta,\tau\rightarrow0)$, with $b$-th entry being one and the remaining entries being 0.
The defined transformation generates $\mr{cumsum}'(\mb{c}^\ast)=[\underbrace{0,\dots,0}_{b}\underbrace{1,\dots,1}_{K-b}]$.
Thus, $\bz^\ast=\mb{e}-\mr{cumsum}'(\mb{c}^\ast)=[\underbrace{1,\dots,1}_{b}\underbrace{0,\dots,0}_{K-b}]$.
It is easy to see the transformation $t'(\cdot)$ is surjective function where  $t': \{\mr{one\_hot}(i)\}_{i=1}^K\rightarrow\mc{V}, t'(\mb{c})=\mb{e}-\mr{cumsum}'(\mb{c})$.
$\mc{V}$ is exactly the set of ordered mask defined in Section~2.2.

Thus, we can calculate the approximation of Downhill variable to the Bernoulli chain,
\begin{equation}
\mr{KL}[q(\bz)||p(\bz)]=\sum_{j=1}^K q(\bv_j)\log\frac{q(\bv_j)}{p(\bv_j)},\label{zapprox}
\end{equation}
where $q(\bv_j) = \beta_j, p(\bv_j) = (1-\pi_{j+1})\prod_{k=1}^j \pi_k$ (See Appx~A.2).
The KL divergence in (\ref{zapprox}) minimized to 0 when $\beta_j=(1-\pi_{j+1})\prod_{k=1}^j, \forall j\in [1,\dots,K]$.

\subsection{Probability of ordered masks}
\label{ap-proof-probability}

Recall the formulation of the Bernoulli chain:
\begin{align}
& p(z_1 = 1)=\pi_1,     & &p(z_1 = 0) =1-\pi_1,
\\
& p(z_i=1|z_{i-1}=1)=\pi_i, & &p(z_i=0|z_{i-1}=1)=1-\pi_i, 
\nonumber\\
& p(z_i=1|z_{i-1}=0)=0, & &p(z_i=0|z_{i-1}=0)=1,
\nonumber
\end{align}

It is observed, there is a chance that $z_i=1$ only when $z_{i-1}=1$. However, if $z_i=0$, then $z_{j}=0$ for $j>i$.
Thus, 
\begin{equation}
p(\bz=\bv_j)=(1-\pi_{j+1})\prod_{k=1}^j\pi_k,
\end{equation}
where $j+1$ is the index of first zero.
For convenience, we define $\pi_{K+1}=0$ as $p(\bz=\bv_K)=\prod_{k=1}^K\pi_k$, which means all nodes are kept.

\subsection{Posterior approximation - $\Phi_2$}
\label{ap-proof-phi2}

Define $\bw_j=[w_{1j},\dots,w_{Dj}]$ as the $j$-th column of $\bW$, and
$q_{\btheta}(\bw_j|z_j=k)=q_{\btheta}(\bw_j|z_j^k)$ where $k\in\{0,1\}$. The term $\Phi_2$ of  (\ref{eq:reg}) is
\begin{align}
	\Phi_2=&\sum_{\bz\in\calV}q_{\bbeta}(\bz)\sum_j\int_{\bw_j} q_{\btheta}(\bw_j|z_j^k)\log\frac{q_{\btheta}(\bw_j|z_j^k)}{p(\bw_j|z_j^k)} \abc{d\bw_j}\nonumber\\
	=&\sum_{\bz\in\calV}q_{\bbeta}(\bz)\sum_i\sum_j\int_{w_{ij}} q_{\btheta}(w_{ij}|z_j^k)\log\frac{q_{\btheta}(w_{ij}|z_j^k)}{p(w_{ij}|z_j^k)} \abc{dw_{ij}}\nonumber\\
	=&\sum_{\bz\in\calV}q_{\bbeta}(\bz)\sum_{i,j} \KL[q_{\btheta}(w_{ij}|z_j^k)||p(w_{ij}|z_j^k)].\label{eq:mlp-kl-2a-ap}
\end{align}
Note that the term inside the integration over $w_{ij}$ is the KL divergence between the univariate conditional density in the prior and the posterior, with $z_j=0$ or $z_j=1$.
%
%
Define $K_{ij}^k(\btheta)$ as the KL of $w_{ij}$ for component $k\in \{0,1\}$. The term $\Phi_2$ can then be re-organized as 
\begin{align}
\begin{split}
	\Phi_2 =& q(\bz=\bv_1)(\sum_i K_{i1}^1(\btheta)+\sum_{j=2}^{D}\sum_i K_{ij}^0(\btheta)) \\
	& +q(\bz=\bv_2)(\sum_{j=1}^2\sum_i K_{ij}^1(\btheta)+\sum_{j=3}^{D}\sum_i K_{ij}^0(\btheta)) \\ 
	& +\dots
	\end{split}
\end{align}
There are totally $D^2 d$ terms, which \abc{potentially} causes a large computation cost in every epoch.
Consider the matrices $\bs{\kappa}^0_{\btheta}=[K_{ij}^0(\btheta)]_{ij}\in\mbb{R}^{d\times D}$ and $\bs{\kappa}^1_{\btheta}=[K_{ij}^1(\btheta)]_{ij}\in\mbb{R}^{d \times D}$, which are easily computed by applying the KL function element-wise.
The term $\Phi_2$ is then expressed as
\begin{align}
\Phi_2 = 
	\mathbf{e}^T \bs{\kappa}^0_{\btheta} (\mb{J}-\mb{J}_L)^T \bs{\beta}+
	\mathbf{e}^T\bs{\kappa}^1_{\btheta} \mb{J}_L^T \bs{\beta},\label{eq:mlp-kl-2b-ap}
\end{align}
where $\mathbf{e}$ is a vector of 1s, $\mb{J}$ is a matrix of 1s and $\mb{J}_L$ is a lower triangular matrix with each element being 1.
Then the calculation in (\ref{eq:mlp-kl-2b-ap}) can be easily parallelize with a modern computation library.
\subsection{$\ell$-0 regularization}
\label{ap-proof-l0}

We consider the case when the prior over each weight is a spike-and-slap distribution, i.e., $p(w_{ij}|z_j=0)=\delta(w_{ij})$ and $p(w_{ij}|z_j=1)=\mc{N}(w_{ij}|0,1)$, using the notation in Section~3.3.
The posterior is also in this form.
The derivations of KL term $\bs{\kappa}$ remain unchanged as it makes nothing but mean-field assumption on the weight prior.
With $\Phi_1$ and $\Phi_2$, the objective (7) can be re-organized as
\begin{align}
\mc{L}_{\btheta,\bbeta}^{\mr{SGVB}}
\simeq &L_{\mc{D}}^{\mr{SGVB}}(\btheta,\bbeta)-\sum_{j=1}^{D} \KL[q_{\bbeta}(\bv_i)||p(\bv_i)]
-\mathbf{e}^T \bs{\kappa}^0_{\btheta} (\mb{J}-\mb{J}_L)^T \bs{\beta}-
\mathbf{e}^T\bs{\kappa}^1_{\btheta} \mb{J}_L^T \bs{\beta}\\
=&L_{\mc{D}}^{\mr{SGVB}}(\btheta,\bbeta)-\sum_{j=1}^{D} \KL[q_{\bbeta}(\bv_i)||p(\bv_i)]
-\mathbf{e}^T\bs{\kappa}^1_{\btheta} \mb{J}_L^T \bs{\beta},\label{eq:obj1}
\end{align}
since $\mr{KL}[q(w_{ij}|z_j=0)||p(w_{ij}|z_j=0)]=0$.
We assume $\mr{KL}[q(w_{ij}|z_j=1)||p(w_{ij}|z_j=1)]=\chi$ as in \cite{louizos2018learning}, which 
means that transforming $p(w_{ij}|z_j=1)$ to $q(w_{ij}|z_j=1)$ requires $\chi$ nats.
Thus, $\mb{K}_{\btheta}^1=[\chi]_{d\times D}$.
The last term is then simplified to
\begin{equation}
	-\chi d \sum_{j=1}^D j\beta_j
\end{equation}
Then,
\begin{align}
	\mc{L}_{\btheta,\bbeta}^{\mr{SGVB}}
	&= L_{\mc{D}}^{\mr{SGVB}}(\btheta,\bbeta)-\sum_{j=1}^{D} \KL[q_{\bbeta}(\bv_i)||p(\bv_i)]
	-\chi d \sum_{j=1}^D j\beta_j,\label{eq:obj2}\\
	&\leq L_{\mc{D}}^{\mr{SGVB}}(\btheta,\bbeta)
	-\chi d \sum_{j=1}^D j\beta_j\label{eq:obj3}
\end{align}
where the inequality is because KL is \abc{non-negative}.
Let $\lambda=\chi d$.
Then, maximizing the evidence lower bound presents the same objective in (16).
This objective assigns greater penalization to the larger sub-networks with more redundant nodes.
To compare with (23)~\cite{louizos2018learning} that uses  a constant coefficient over the probabilities, our reduced formulation provides an \emph{ordered $\ell$-0 regularization} instead of a \emph{uniform $\ell$-0 regularization}.

Note that (\ref{eq:obj2}) ignores the weight uncertainty compared with (7).
(\ref{eq:obj3}) further ignores the uncertainty over the ordered mask, reduced to a deterministic formulation for a \emph{nested neural network with learned weight importance}.

\subsection{Discussion of Regularization}
\label{appx:elbo}
The vanilla variational auto-encoder suffers from the problem of posterior collapse.
During optimization, the KL term could reduce to 0, thus the approximate posterior equals the prior, which indicates no information is learned from the data.
However, a useful approximate inference requires the KL to be positive.
\yedit{Previously, $\delta$-VAE}~\cite{razavi2019preventing} proposes autoregressive latent variables to enforce the KL term to be positive.
%
\yedit{Our method has a simple but effective structure over the latent space, and is compatible to the previous advances.}

We assume a hard sample obtained from the Downhill distribution, and 
%
denote $\kappa_i$ as the KL divergence calculated for variable $h_i$. 
\yedit{According to the parameterization in Section~\ref{sec:vnd_ov}, the KL term is
\begin{align}
\begin{split} &\KL[q(\bh,\bz|\bx)||p(\bh,\bz)] \\
&=\mathbb{E}_{q(\mathbf{z})}[\log\frac{q(\mathbf{z})}{p(\mathbf{z})}]
+\mathbb{E}_{q(\mb{z})}\underbrace{\mathbb{E}_{q(\bh|\mb{z})}[\log\frac{q(\bh|\mb{z})}{p(\bh|\mb{z})}]}_{\bs{\kappa}=[\kappa_i]_i} \\
&=\underbrace{\sum_i \beta_i\log\frac{\beta_i}{(1-\pi_{i+1})\prod_{k=1}^i\pi_k}}_{\Phi_1}+\underbrace{\sum_i\beta_i\kappa_i}_{\Phi_2}
\end{split}
\label{eq:postKL1}
\end{align}
Given a fixed $[\kappa_i]_i$, we solve for $[\beta_i]_i$ to understand the structure that VND brings to the regularization term. 
We could massage $\Phi_2$ to such that
\begin{align}
\Phi_2 = \sum_i \beta_i \kappa_i = -\sum_i \beta_i \log e^{-\kappa_i}
\end{align}
The KL divergence term could be combined as
\begin{align}
\begin{split}
\KL &= \Phi_1+\Phi_2 \\
&= \sum_i \beta_i \log \frac{\beta_i}{e^{-\kappa_i}(1-\pi_{i+1})\prod_{k=1}^i\pi_k}
\end{split}
\end{align}
To make it a proper KL for the Gaussian-Bernoulli mixture, we normalize the second distribution with the normalizing constant
$C = \sum_i [e^{-\kappa_i}(1-\pi_{i+1})\prod_{k=1}^i\pi_k]$.
\begin{align}
\begin{split}
\label{eq:trans_kl}
\KL &= \sum_i \beta_i \log  \frac{\beta_i}{e^{-\kappa_i}(1-\pi_{i+1})\prod_{k=1}^i\pi_k}\frac{C}{C}\\
&= \sum_i \beta_i \log \frac{\beta_i}{e^{-\kappa_i}(1-\pi_{i+1})\prod_{k=1}^i\pi_k/C} - \sum_i \beta_i \log C\\
&= \sum_i \beta_i \log \frac{\beta_i}{e^{-\kappa_i}(1-\pi_{i+1})\prod_{k=1}^i\pi_k/C} - \log C
\end{split}
\end{align}
The first term is now a proper KL divergence. 
The minimum value of 0 occurs when $\beta_i = e^{-\kappa_i}(1-\pi_{i+1})\prod_{k=1}^i\pi_k / C$, then $\KL = -\log C$.
In this case, if $\kappa_i=0,\forall i$, then $C=1$ and $\KL=0$.
If $\kappa_i>0$, then $C<1$ and $\KL>0$.
If $\kappa_i$ is lower bounded with any previous advance, the $\KL$ is lower bounded, such that the mode collapse could be avoided.
}

\yedit{\textbf{Diversity:} As discussed in Section~\ref{sec:divers}, as long as the posterior does not collapse to the \emph{single-modal} case, the diversity could be guaranteed.
The \emph{single-modal} case corresponds to $\beta_1=1$ and $\beta_i=0, \forall i>1$. The form of KL in (\ref{eq:trans_kl}) prevents such case. 
Consider the situation when the optimal $\beta_i$ is reached:
\begin{itemize}
    \item If $\kappa_i=0,\forall i$, then $C=1$ and $\KL=0$. The posterior collapses to the prior (this could be avoided as discussed above). $\beta_i = (1-\pi_{i+1})\prod_{k=1}^i\pi_k$, still keeps a geometric distribution form shown in the left graph in Fig.~\ref{fig:motivation} (Rippel~\emph{et al}). The multi-modal structure is kept.
    \item If $\kappa_i>0$, $\beta_i = e^{-\kappa_i}(1-\pi_{i+1})\prod_{k=1}^i\pi_k / C > 0$. The single model case is avoided.
\end{itemize}
}

\section{Implementation}

\subsection{Extension to Convolutional Layer}
\label{ap-impl-conv}
We consider a convolutional layer takes in a single tensor $\mathbf{H}_m^{H\times W\times C}$ as input, where $m$ is the index of the batch, $H$, $W$ and $C$ are the dimensions of feature map.
The layer has $D$ filters aggregated as $\bw^{D\times H'\times W'\times C}$ and outputs a matrix $\mb{F}_{mj}^{\bar{H}\times\bar{W}}$.
In the paper, we consider the ordered masks applied over the output channels and each filter corresponds to a dimension in $\bz$.
As shown in~\cite{kingma2015variational,molchanov2017variational}, the local reparameterization trick can be applied, due to the linearity of the convolutional layer.
\begin{align}
	&f_{mj}=b_{mj}z_j^\ast,\quad \mr{vec}(b_{mj})\sim \mc{N}(\gamma_{mj},\delta_{mj}) \label{sampling-conv}\\
	&\gamma_{mj}=\mr{vec}(\mb{H}_m\ast\bw),\quad \delta_{mj}=\mr{diag}(\mr{vec}(\mb{H}_m^2\ast\sigma_j^2))\nonumber
\end{align}
where $z_j^\ast$ is the $j$-th dimension of the sampled ordered mask $\bz^\ast=\bv^\ast \sim q_{\bbeta}(\bz)$.

To calculate the KL term (11), the only modification is to let the first summation be over the height, width and input channels in (13).
\begin{equation}
	\Phi_2=\sum_{\bz\in\calV}q_{\bbeta}(\bz)\sum_i^{H'\times W'\times C}\sum_j^D\int_{w_{ij}} q_{\btheta}(w_{ij}|z_j^k)\log\frac{q_{\btheta}(w_{ij}|z_j^k)}{p(w_{ij}|z_j^k)}
\end{equation}

\subsection{Re-scale weights for testing}
\label{ap-exp-rescaling}
During training, the network drops nodes with the variational nested dropout.
In testing, the network fixes width of each layer and no dropout operation is adopted.
To make the expectation consistent over training and testing~\cite{srivastava2014dropout}, we re-scale the weights according to the probability to keep a node.
\begin{equation}
	\mbb{E}_{\bz\sim q_\beta(\bz), \bx\sim \mc{D}_{\mr{tr}}}[\mb{F}|\bx,\bz] \approx \mbb{E}_{\bx\sim \mc{D}_{\mr{te}}}[\mb{F}|\bx,\bz=\bar{\bv}],
\end{equation}
where $\mc{D}_{\mr{tr}}$ and $\mc{D}_{\mr{te}}$ are the splits of training set and testing set, and $\bar{\bv}$ is the user-specified width according to the real demand during testing time.

We take the fully-connected layer as an example.
For simplicity, we treat $w_{ij}$ as deterministic here.
\begin{align}
	\mbb{E}_{\bz\sim q_\beta(\bz)}[f_{mj}]
	=\mbb{E}_{\bz\sim q_\beta(\bz)}\big[z_j\sum_{i=1}^d h_{mi}\theta_{ij}\big]
	=\mbb{E}_{\bz\sim q_\beta(\bz)}[z_j]\sum_{i=1}^d h_{mi}\theta_{ij}\label{eq:rescale}
\end{align}
Note that, different from the probability $q_\beta(\bv_j)=\beta_j$, $\mbb{E}_{\bz\sim q_\beta(\bz)}[z_j]$ is the probability that the $j$-th node is kept.

With a well-trained layer in a Bayesian nested neural network, we have the learned importance $\bbeta=[\beta_j]_j$.
Assume that the $\bbeta$ is also generated by a chain of hidden Bernoulli variables following (1) with the parameters $\bs{\mu}=[\mu_j]_j$, with $\mu_1\coloneqq 1$ and $\mu_j=q(z_j=1|z_{j-1}=1)$.
We are interested in the marginal distribution $p(z_j=1)=\prod_{k=1}^j\mu_k$ but we only have $\beta_j$'s.
\begin{align}
	&\beta_1=(1-\mu_2)\mu_1=1-\mu_2\\
	&\beta_2=(1-\mu_3)\mu_2\mu_1\nonumber\\
	&\dots\nonumber
\end{align}
Solving each equation sequentially, we obtain 
\begin{align}
	&p(z_1=1)=1,\nonumber\\
	&p(z_2=1)=1-\beta_1,\nonumber\\
	&p(z_3=1)=1-\beta_1-\beta_2,\nonumber\\
	&\dots
\end{align}
and (\ref{eq:rescale}) becomes
\begin{equation}
	\abc{\mbb{E}_{\bz\sim q_\beta(\bz)}[f_{mj}]} = 	(1-\sum_{k=1}^{j-1}\beta_k)\sum_{i=1}^d h_{mi}\theta_{ij},
\end{equation}
where we can define $\beta_0=0$.
Then, the \emph{scaling factor} is $1-\sum_{k=1}^{j-1}\beta_k$ for each $w_{ij}$.

Another way is to optimize the conditional probabilities $[\mu_j]_j$ instead of $\beta_j$, with $\beta_j$ in previous derivation replaced by $(1-\mu_{j+1})\mu_j$\footnote{We use this parameterization in our implementation, while we use $\beta_j$ in most of our derivation for simplicity in writing.}.
The scaling factor is then $\prod_{k=1}^j\mu_k$ for each $w_{ij}$.
Also, for simplicity, one can optimize $\bar{\mu}_k$ where $\mu_k=\mr{sigmoid}(\bar{\mu}_k)$.

\section{Experiments}
\subsection{Experimental setups}
\label{ap-exp-setup}
We implement \FNNN, individual Bayesian neural networks (IBNN) and the proposed Bayesian Nested Neural Network (\BNNN) with PyTorch framework.
We use the cross-entropy loss for negative expected log-likelihood.
For balancing the regularization and likelihood, we add a scaling factor $\kappa$ for the KL term, which is a common trick in Bayesian learning~\cite{higgins2016beta}.

\subsubsection{Cifar10/Cifar100.} For data augmentation, we use random cropping with padding beforehand, and random flipping the image horizontally.

\emph{VGG11}: We train \BNNN-VGG11 with natural gradient descent\footnote{The PyTorch implementation is from https://github.com/YiwenShaoStephen/NGD-SGD.} (NGD)~\cite{povey2014parallel}, as it was shown to make the Bayesian neural network converge faster~\cite{khan2018fast}.
The network is trained for 600 epochs with an initial learning rate 0.1 and momentum 0.9.
The learning rate is scaled by a factor 0.1 every 150 epochs.
$\kappa$ is set to $10^{-5}$.
For training the network, we use VGG11 with $1.5\times$ number of channels and truncate the $2/3$ part with higher importance for testing.
We add one dense layer after the stack of convolutional layers.
The first feature extraction layer and the last two dense layers for classification are variational Bayes layer without nested dropout, with our parameterization proposed in Section~3.4.
For the convolutional layer, we divide the convolutional filters into 32 groups for group sparsity.
30 groups are applied nested dropout while the remaining 2 groups are for extracting the basic features.
The $\log\alpha_{ij}$ is initialized to -8 for the first layer and -1 for the rest layers.
The $[\bar{\mu}_j]_j$ are all initialized to 3.
We train IBNN-VGG11 with NGD for 240 epochs, with an initial learning rate 0.1 and scaled by 0.3 every 40 epochs.
Every individual BNN is fixed at some width between the fraction 0 and 1.
We train \FNNN-VGG11 with SGD and momentum 0.9, as SGD performs better in training \FNNN-VGG11.
Other setups are similar to \BNNN-VGG11.

\emph{MobileNetV2}: We train \BNNN-MobileNetV2 with a similar setup as \BNNN-VGG11, except the following.
For inverted residual block, we apply nested dropout to the middle depth-wise convolutional layer, 
because it already sparsifies the convolution filters in the previous point-wise convolution layer, and channels in the following point-wise convolution layer~\cite{he2017channel}.
Introducing more nested dropout units would cause extra and irregular sparsification which deteriorates the performance.
We use a normal-size MobileNetV2 and divide the weights into 16 groups.
One group is fixed for base feature extraction.
The experimental setups for IBNN-MobileNetV2 and \FNNN-MobileNetV2 follow that on VGG11.

\emph{ResNeXt-Cifar}: The setups for ResNeXt-Cifar are similar to that of MobileNetV2, while the number of groups is 32.

\subsubsection{Tiny-ImageNet.} 
For data augmentation, we use random cropping with padding beforehand, random rotation of 20 degree and random flipping the image horizontally.
All images are finally cropped to $64\times64$ and all networks are trained from scratch.

\emph{VGG11}: To increase the capacity, we take VGG11 with $1.5\times$ number of channels as the base network.
The network is trained with NGD for 300 epochs with an initial learning rate 0.1.
The learning rate is scaled by 0.3 every 25 epochs.
$\kappa$ is set to $10^{-6}$.
The weights are divided into 32 groups and 8 groups are fixed for base feature extraction.

\emph{MobileNetV2}: We train a MobileNetV2 with $1.5\times$ number of channels, and take the $2/3$ part with higher importance as the base network for testing.
The network is trained with NGD for 300 epochs with an initial learning rate 0.1.
The learning rate is scaled by 0.3 every 40 epochs.
The weights are divided into 16 groups and 1 groups are fixed for base feature extraction.

\emph{ResNeXt-Cifar}: We train a ResNeXt-Cifar with normal size.
The weights are divided into 32 groups and 8 groups are fixed for base feature extraction.
The network is trained with SGD for 300 epochs with an initial learning rate 0.1.
The learning rate is scaled by 0.3 every 30 epochs.

The remaining setups are similar to that on Cifar10/Cifar100.

\subsubsection{Lung Abnormalities Segmentation.} 

The network uses a UNet shape architecture with layers 32-64-128-192 for the encoder (two layers fewer than the standard UNet).
The optimizer is Adam with initial learning rate $10^{-4}$ decayed by 0.1 every 60 epochs.
The channels are divided into 32 groups and 6 groups are fixed for base feature extraction.

\subsection{BN statistics}
\label{ap-exp-bn}
\begin{figure}[t]
	\begin{center}
	\includegraphics[width=0.6\linewidth]{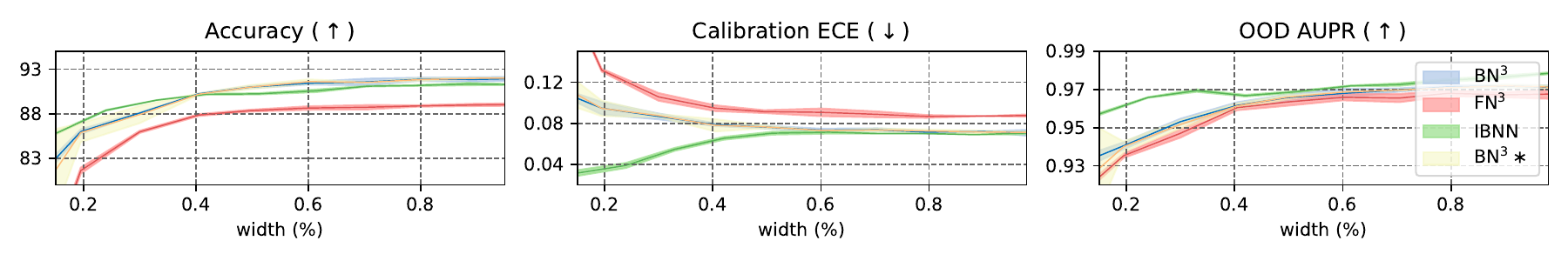}
	\end{center}
	\caption{Performance of VGG11 on Cifar10 with less data for collecting BN statistics.} 
	\label{fig:bn}
\end{figure}

We show that collecting batch normalization statistics on a small training set yields similar performance to using the whole dataset
In this example, we use VGG11 on Cifar10.
The collection proceeds by forwarding the network by 2 iterations, with a batch size 512.
Thus, in total, $1024/50000$ training data are used for statistics collection.
The results are shown in Figure~\ref{fig:bn} as $\mr{BN}^3\ast$.
We can observe that this results are similar to using all training data for statistics collection, with slightly larger variance using a lower width.

\begin{figure}[t]
	\begin{center}
		\includegraphics[width=0.6\linewidth]{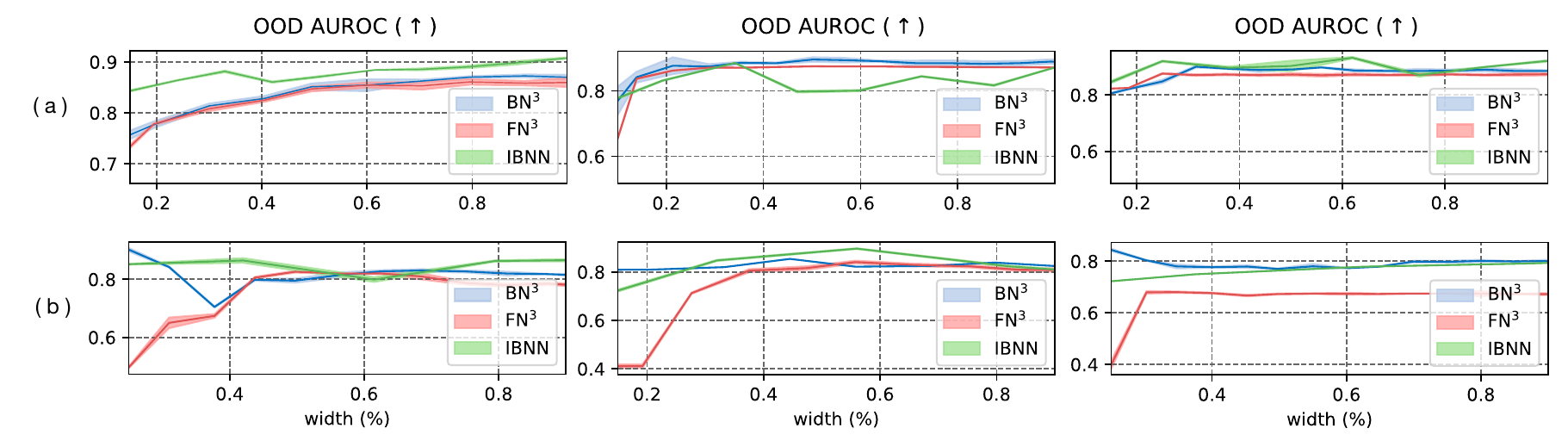}
	\end{center}
	\caption{The AUROC of OOD on (a) Cifar10 (b) Tiny ImageNet datasets with VGG11, MobileNetV2 and ResNeXt-Cifar (left to right).} 
	\label{fig:ood}
\end{figure}

\subsection{OOD detection}
\label{ap-exp-ood}
For out-of-domain detection, we use the SVHN dataset as the OOD data~\footnote{\url{http://ufldl.stanford.edu/housenumbers/}}.
The OOD detection performance with AUROC metric is shown in Figure~\ref{fig:ood}.
The performance is similar to that of AUPR in Figure 5.

\subsection{Cifar100 results}
\label{ap-exp-cifar10}
\begin{figure}[hb]
	\begin{center}
		\includegraphics[width=0.6\linewidth]{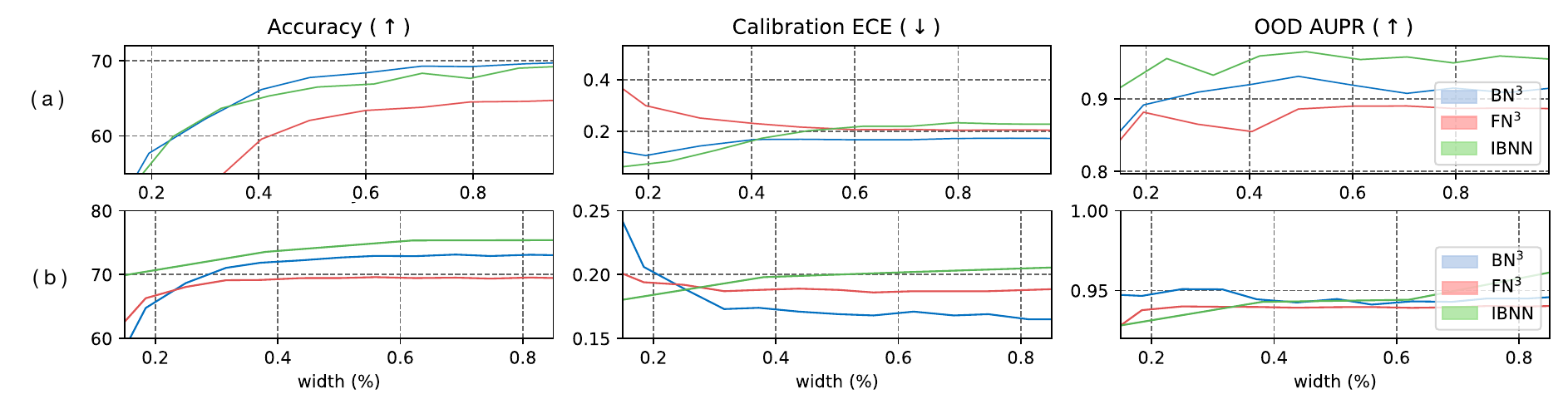}
	\end{center}
	\caption{Results on Cifar100 for (a) VGG11, (b) ResNeXt-Cifar.} 
	\label{fig:c100}
\end{figure}

The results on Cifar100 is shown in Figure~\ref{fig:c100}.
As the hyper-parameters are mostly from training on Cifar10, the results may not be optimal.
We do not show the comparisons for MobileNetV2 here, as it is observed that IBNN-MobileNetV2 fails provide a decent performance on Cifar100, similar to Figure 5(b).
The proposed \BNNN performs well steadily on every task.

\end{document}


\title{\LaTeX\ Author Guidelines for CVPR Proceedings}

\author{First Author\\
Institution1\\
Institution1 address\\
{\tt\small firstauthor@i1.org}
\and
Second Author\\
Institution2\\
First line of institution2 address\\
{\tt\small secondauthor@i2.org}
}

\maketitle


\section{Proofs}
\subsection{Downhill random variables}
We assume $\mathbf{c}$ follows a Gumbel softmax distribution~\cite{gumbel1948statistical, maddison2014sampling} which has the following form.
\begin{align}
&p(c_1,\dots,c_K)\\
=&\Gamma(K)\tau^{K-1}(\sum_{i=1}^K \pi_i/c^\tau)^{-K}\prod_{i=1}^K(\pi_i/c^{\tau+1})\nonumber
\end{align}
We apply the transformation $T_i(\cdot)=\mathbf{e}_i-\mathrm{cumsum}_i^\prime(\mathbf{\cdot})$ to the variable $\mathbf{c}$.
$\mathbf{z}=T(\mathbf{c})=\mathbf{e}-\mathrm{cumsum}_i^\prime(\mathbf{c})$

To obtain the distribution of $p(\mathbf{z})$, we apply the change of variables formula on $\mathbf{c}$.
\begin{align}
	p(\mathbf{z})=p(T^{-1}(\mathbf{z}))|\mathrm{det}(\partial \frac{T^{-1}(\mathbf{z})}{\partial \mathbf{z}})|\\
	p(z_{1:K})=p(T^{-1}(z_{1:K}))|\mathrm{det}( \frac{\partial T^{-1}(z_{1:K})}{\partial z_{1:K}})|
\end{align}
From the definition of $T(\cdot)$, we can obtain $T_i^{-1}(\mathbf{z})=z_{i-1}-z_i$.
The Jacobian
\begin{align}
	\frac{\partial T^{-1}(z_{1:K})}{\partial z_{1:K}}=
	\begin{bmatrix}
	-1 & 0  & \dots & 0 & 0 \\
	1  & -1 & \dots & 0 & 0 \\
	0  & 1  & \dots & 0 & 0 \\
	\vdots & & & & \vdots \\
	0  & 0  & \dots & 1 & -1
	\end{bmatrix}
\end{align}
Thus, $|\mathrm{det}(\frac{\partial T^{-1}(z_{1:K})}{\partial z_{1:K}})|=1$.

\begin{align}
&p(z_{1:K})=p(T_{1:K}^{-1}(\mathbf{z}))\\
=&\Gamma(K)\tau^{K-1}(\sum_{i=1}^K\frac{ \pi_i}{(z_{i-1}-z_i)^\tau})^{-K}\prod_{i=1}^K(\frac{\pi_i}{(z_{i-1}-z_i)^{\tau+1}})\nonumber
\end{align}


{\small
\bibliographystyle{ieee_fullname}
\bibliography{egbib}
}